\documentclass[letterpaper]{article} 
\usepackage{aaai2026}  
\usepackage{times}  
\usepackage{helvet}  
\usepackage{courier}  
\usepackage[hyphens]{url}  
\usepackage{graphicx} 
\usepackage{natbib}  
\usepackage{caption} 
\usepackage{algorithm}
\usepackage{algorithmic}

\usepackage{bbding}
\usepackage{xcolor}
\usepackage{amsmath,amssymb}
\usepackage{subcaption}
\usepackage{booktabs,multirow,makecell}
\usepackage[table]{xcolor}

\usepackage{newfloat}
\usepackage{listings}
\DeclareCaptionStyle{ruled}{labelfont=normalfont,labelsep=colon,strut=off} 
\floatstyle{ruled}
\newfloat{listing}{tb}{lst}{}
\floatname{listing}{Listing}
\title{Transparent Networks for Multivariate Time Series}
\author{
    Minkyu Kim\textsuperscript{\rm 1},
    Suan Lee\textsuperscript{\rm 2}\thanks{Corresponding author.},
    Jinho Kim\textsuperscript{\rm 3,\rm 4}\footnotemark[1]
}
\affiliations{
    \textsuperscript{\rm 1}Ziovision Co., Ltd.\\
    \textsuperscript{\rm 2}Semyung University\\
    \textsuperscript{\rm 3}Kangwon National University\\
    \textsuperscript{\rm 4}Korea Information Systems Consulting \& Audit Co., Ltd.\\
    minkyu.kim@ziovision.co.kr, suanlee@semyung.ac.kr, jhkim@kangwon.ac.kr
}

\begin{document}

\maketitle

\begin{abstract}
Transparent models, which provide inherently interpretable predictions, are receiving significant attention in high-stakes domains. However, despite much real-world data being collected as time series, there is a lack of studies on transparent time series models. To address this gap, we propose a novel transparent neural network model for time series called Generalized Additive Time Series Model (GATSM). GATSM consists of two parts: 1) independent feature networks to learn feature representations, and 2) a transparent temporal module to learn temporal patterns across different time steps using the feature representations. This structure allows GATSM to effectively capture temporal patterns and handle varying-length time series while preserving transparency. Empirical experiments show that GATSM significantly outperforms existing generalized additive models and achieves comparable performance to black-box time series models, such as recurrent neural networks and Transformer. In addition, we demonstrate that GATSM finds interesting patterns in time series.
\end{abstract}

\begin{links}
    \link{Code}{https://github.com/gim4855744/GATSM}
    \link{Extended version}{https://arxiv.org/abs/2410.10535}
\end{links}

\section{Introduction}

Artificial neural networks excel at learning complex representations and demonstrate remarkable predictive performance across various fields. However, their complexity makes interpreting the decision-making processes of neural network models challenging. Consequently, researchers have widely studied post-hoc explainable artificial intelligence (XAI) methods, which explain the predictions of trained black-box models, in recent years \cite{ribeiro_why_2016,lundberg_unified_2017,selvaraju_grad-cam_2017,mothilal_explaining_2020}. XAI methods are generally effective at providing humans with understandable explanations of model predictions. However, they may produce incorrect and unfaithful explanations of the underlying black-box model and cannot provide actual contributions of input features to model predictions \cite{cynthia_rudin_please_2018,cynthia_rudin_stop_2019,rahnama_study_2019,liu_synthetic_2021,akhavan_rahnama_blame_2023,rahnama_can_2024}. Therefore, their applicability to high-stakes domains -- such as healthcare and fraud detection, where faithfulness to the underlying model and actual contributions of features are important -- is limited.

Due to these limitations, transparent (i.e., inherently interpretable) models are attracting attention as alternatives to XAI in high-stakes domains \cite{rishabh_agarwal_neural_2021,chun-hao_chang_node-gam_2022,filip_radenovic_neural_2022}. Modern transparent models typically adhere to the \textit{generalized additive model} (GAM) framework \cite{hastie_generalized_1986}. A GAM consists of independent functions, each corresponding to an input feature, and makes predictions as a linear combination of these functions (e.g., the sum of all functions). Therefore, each function reflects the contribution of its respective feature. For this reason, interpreting GAMs is straightforward, making them widely used in various fields, such as healthcare \cite{rich_caruana_intelligible_2015,chang_how_2021}, survival analysis \cite{utkin_survnam_2022}, and model bias discovery \cite{rishabh_agarwal_neural_2021,tan_distill-and-compare_2018,kim_higher-order_2022}. However, despite much real-world data being collected as time series, research on GAMs for time series remains scarce. Consequently, the applicability of GAMs in real-world scenarios is still limited.

To overcome this limitation, we propose a novel transparent model for multivariate time series called Generalized Additive Time Series Model (GATSM). GATSM consists of independent feature networks to learn feature representations and a transparent temporal module to learn temporal patterns. Since employing distinct networks across different time steps requires a massive amount of learnable parameters, the feature networks in GATSM share the weights across all time steps, while the temporal module learns temporal patterns. GATSM then generates final predictions by integrating the feature representations with the temporal information from the temporal module. This strategy allows GATSM to effectively capture temporal patterns and handle dynamic-length time series while preserving transparency. Additionally, this approach facilitates the separate extraction of time-independent feature contributions, the importance of individual time steps, and time-dependent feature contributions through the feature functions, temporal module, and final prediction. To demonstrate the effectiveness of GATSM, we conducted empirical experiments on various real-world and synthetic time series datasets. The experimental results show that GATSM significantly outperforms existing GAMs and achieves comparable performance to black-box time series models, such as recurrent neural networks and Transformer \cite{ashish_vaswani_attention_2017}. In addition, we provide visualizations of GATSM's predictions to demonstrate that GATSM finds interesting patterns in time series.

A detailed discussion of related works is provided in Appendix~\ref{sec:related_works} and we summarize the contributions of this paper as follows: 1) We propose a novel transparent neural network model for multivariate time series, called GATSM. To the best of our knowledge, GATSM is the first transparent model capable of capturing temporal patterns while preserving transparency. 2) We conduct extensive experiments to validate the performance of GATSM. The experimental results demonstrate that GATSM significantly outperforms existing transparent models and achieves performance comparable to black-box models like Transformer. 3) We provide visual interpretations of GATSM’s predictions. These visualizations illustrate that GATSM successfully captures complex temporal patterns and derive multiple forms of interpretations, including time-step importance, global feature contributions, local time-independent feature contributions, and local time-dependent feature contributions. Such interpretations significantly enhance the understanding of both model behaviors and underlying dataset characteristics. 4) We thoroughly discuss the potential applications and limitations of GATSM, identifying directions for future works.

\section{Problem Statement}

The simple linear model is designed to fit the conditional expectation $g \left( \mathbb{E} \left( y \mid \textbf{x} \right) \right) = \sum_{i=1}^{M} x_i w_i$, where $g(\cdot)$ is a link function, $M$ indicates the number of input features, $y$ is the target value for the given input features $\textbf{x} \in \mathbb{R}^{M}$, and $w_i \in \mathbb{R}$ is the learnable weight for $x_i$. This model captures only linear relationships between the target $y$ and the inputs $\textbf{x}$. To address this limitation, GAM \cite{hastie_generalized_1986} extends the simple linear model to the generalized form as follows:
\begin{equation}
    g \left( \mathbb{E} \left( y \mid \textbf{x} \right) \right) = \sum_{i=1}^M f_i \left( x_i \right),
    \label{eq:gam}
\end{equation}
where each $f_i(\cdot)$ is a function that models the effect of a single feature, referred as a feature function. Typically, $f_i \left( \cdot \right)$ becomes a non-linear function such as a decision tree or neural network to capture non-linear relationships. However, this form of GAM cannot handle time series data. One straightforward method to extend GAM to time series, adopted in NATM \cite{jo_neural_2023}, is applying distinct feature functions to each time step and summing them to produce predictions:
\begin{equation}
    g \left( \mathbb{E} \left( y_t \mid \textbf{X}_{:t} \right) \right) = \sum_{i=1}^{t} \sum_{j=1}^{M} f_{i,j} \left( x_{i,j} \right),
\end{equation}
where $\textbf{X} \in \mathbb{R}^{T \times M}$ is a time series with $T$ time steps and $M$ features, and $t$ is the current time step. This method can handle time series data as input but fails to capture temporal patterns because $f_{i,j} \left( \cdot \right)$ still does not interact with previous time steps. To overcome this problem, we suggest a new form of GAM for time series.

\textbf{Definition 3.1} \textit{GAMs for time series, which capture temporal patterns hold the following form}:
\begin{equation}
    g \left( \mathbb{E} \left( y_t \mid \textbf{X}_{:t} \right) \right) = \sum_{i=1}^{t} \sum_{j=1}^{M} f_{i,j} \left( x_{i,j}, \textbf{X}_{:t} \right).
    \label{eq:gamts}
\end{equation}

In Equation~(\ref{eq:gamts}), $f_{i,j} \left( \cdot, \cdot \right)$ can capture interactions between current and previous time steps. Therefore, GAMs adhering to Definition 3.1 can capture temporal patterns. However, implementing such a model while maintaining transparency poses challenges. In the following section, we will describe our approach to implementing a GAM that satisfies Definition 3.1. To the best of our knowledge, no existing literature addresses Definition 3.1. Table~\ref{tab:advantages} shows the advantages of our GATSM compared to existing GAMs.

\begin{table}[!t]
    
    \centering

    \resizebox{\linewidth}{!}{
    
        \begin{tabular}{c|c|c|c}

            \toprule
        
             & Time series & Temporal pattern & Dynamic length \\

             \midrule
             
             Tabular GAMs & \color{red}{\XSolidBrush} & \color{red}{\XSolidBrush} & \color{red}{\XSolidBrush} \\
             NATM & \color{teal}{\CheckmarkBold} & \color{red}{\XSolidBrush} & \color{red}{\XSolidBrush} \\
             GATSM (our) & \color{teal}{\CheckmarkBold} & \color{teal}{\CheckmarkBold} & \color{teal}{\CheckmarkBold} \\

             \bottomrule
             
        \end{tabular}

    }


    \caption{Advantages of GATSM. Unlike existing tabular GAMs that cannot handle time series data, and NATM which only supports fixed-length time series without capturing temporal dynamics, GATSM effectively handles varying-length time series and captures complex temporal patterns.}
    
    \label{tab:advantages}
    
\end{table}

\begin{figure*}[!t]
    \centering
    \includegraphics[width=\linewidth]{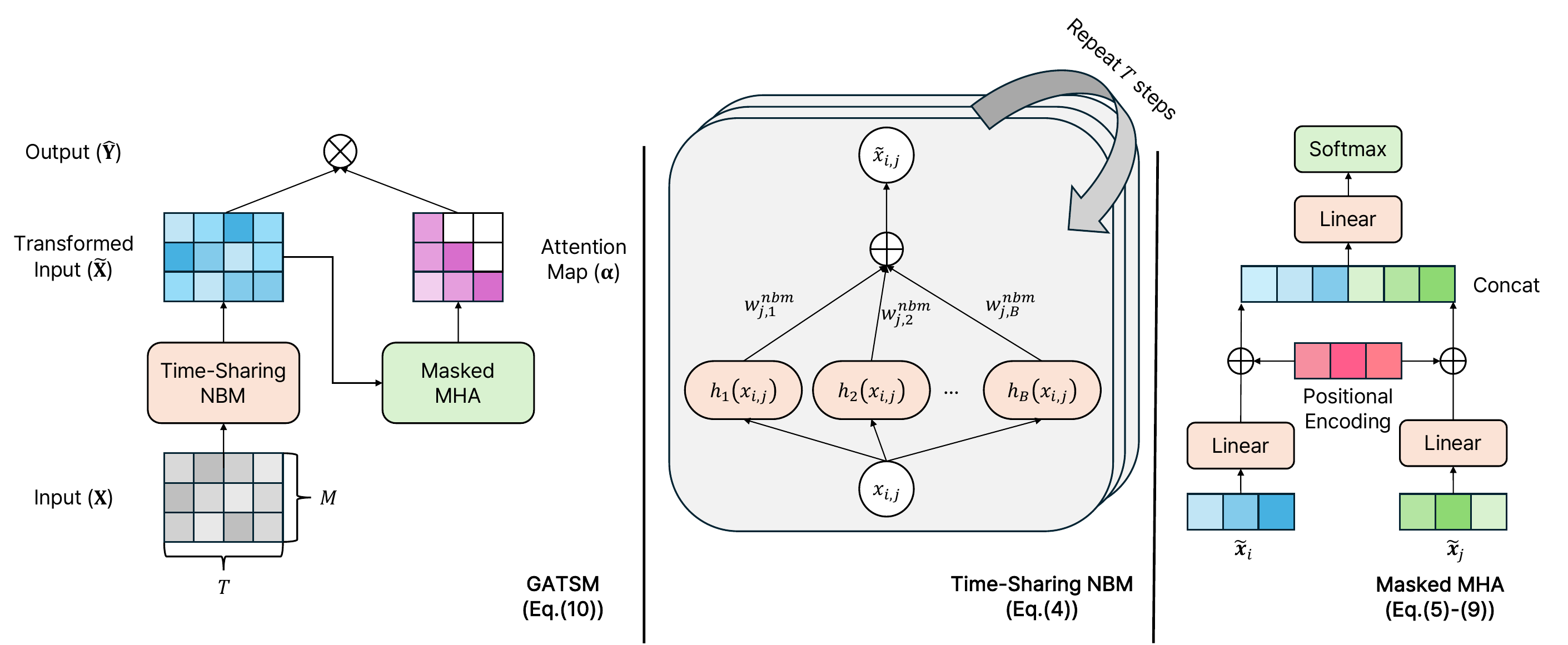}
    \caption{Architecture of GATSM. GATSM comprises two components: Time-Sharing NBM and Masked MHA. Time-Sharing NBM learns nonlinear representations of input features without considering temporal patterns, using parameters shared across time steps. Masked MHA then captures temporal patterns based on these feature representations.}
    \label{fig:gatsm}
\end{figure*}

\section{Generalized Additive Time Series Model}

Figure~\ref{fig:gatsm} shows the overall architecture of GATSM. Our model has two modules: 1) feature networks, called time-sharing neural basis model (NBM), for learning feature representations, and 2) masked multi-head attention (MHA) for learning temporal patterns.

\subsection{Time-Sharing NBM}
Assume a time series $\textbf{X} \in \mathbb{R}^{T \times M}$ with $T$ time steps and $M$ features. Applying existing GAMs defined in Equation~\ref{eq:gam} to this time series require $T \times M$ feature functions, which becomes problematic when dealing with large $T$ or $M$ due to increased model size. This limits the applicability of GAMs to real-world datasets. To overcome this problem, we extend NBM \cite{filip_radenovic_neural_2022} to time series as:
\begin{equation}
    \tilde{x}_{i,j} = f_j \left( x_{i,j} \right) = \sum_{k=1}^{B} h_k \left( x_{i,j} \right) w_{j,k}^{nbm}.
    \label{eq:nbm}
\end{equation}
We refer to this extended version of NBM as time-sharing NBM. Time-sharing NBM has $B$ basis functions, with each basis function $h_k(\cdot)$ taking a feature $x_{i,j}$ as input. The feature-specific weight $w_{j,k}^{nbm}$ then projects the basis to the transformed feature $\tilde{x}_{i,j}$. As Equation~(\ref{eq:nbm}) depicts, the basis functions are shared across all features and time steps, drastically reducing the number of required feature functions from $T \times M$ to $B$. We use $B=100$ and implement $h_k \left( \cdot \right)$ using multi-layer perceptron (MLP).

\subsection{Masked MHA}\label{sec:mha}
GATSM employs MHA to learn temporal patterns. Although the dot product attention \cite{ashish_vaswani_attention_2017} is popular, the simple dot product attention has low expressive power \cite{velickovic_graph_2018}. Therefore, we adopt the 2-layer attention mechanism proposed by \cite{velickovic_graph_2018} to GATSM. We first transform $\tilde{\textbf{x}}_i = \left[ \tilde{x}_{i,1}, \tilde{x}_{i,2}, \cdots, \tilde{x}_{i,M} \right] \in \mathbb{R}^{M}$ produced by Equation~(\ref{eq:nbm}) as follows:
\begin{equation}
    \textbf{v}_i = \tilde{\textbf{x}}_i^{\intercal} \textbf{Z} + \textbf{pe}_i,
    \label{eq:v}
\end{equation}
where $\textbf{Z} \in \mathbb{R}^{M \times D}$ is a learnable weight, $\textbf{pe}_i = \left[ pe_{i,1}, pe_{i,2}, \cdots, pe_{i,D} \right] \in \mathbb{R}^D$ is the positional encoding for the $i$-th step, and $D$ indicates the hidden size. The positional encoding is defined as follows:
\begin{equation}
    pe_{i,j} = 
    \begin{cases}
        \text{sin} \left( \frac{i}{10000^{2j / D}} \right) & \text{if } j \text{ mod } 2 = 1, \\
        \text{cos} \left( \frac{i}{10000^{2j / D}} \right) & \text{otherwise}.
    \end{cases}
    \label{eq:pe}
\end{equation}
The positional encoding helps GATSM effectively capture temporal patterns. While learnable position embedding also works in GATSM, we recommend the sinusoidal positional encoding because learnable position embedding requires prior knowledge of the maximum length of a time series, which is often unknown in real-world settings. For example, in real-world hospital settings, a patient’s length of stay continuously increases, and the discharge date is unpredictable. After computing $\textbf{v}_i$, we calculate the attention scores as follows:
\begin{gather}
    e_{k,i,j} = \sigma \left( \left[ \textbf{v}_i \mid \textbf{v}_j \right]^\intercal \textbf{w}_{k}^{attn} \right) m_{i,j}, \\
    a_{k,i,j} = \frac{\text{exp} \left( e_{k,i,j} \right)}{\sum_{t=1}^{T} \text{exp} \left( e_{k,i,t} \right)},
    \label{eq:mha}
\end{gather}
where $k$ is the index of the attention head, $\sigma \left( \cdot \right)$ is a non-linear activation function, $\textbf{w}_k^{attn} \in \mathbb{R}^{2D}$ is a learnable weight, and $m_{i,j} \in \mathbb{R}$ is the mask value used to block future information. The time mask is defined as follows:
\begin{equation}
    m_{i,j} =
    \begin{cases}
        1 & \text{if } i \geq j, \\
        -\infty & \text{otherwise}.
    \end{cases}
    \label{eq:mask}
\end{equation}

\subsection{Inference}
GATSM produces its prediction by combining the transformed features from the time-sharing NBM with the attention scores from the masked MHA.
\begin{equation}
    \hat{y}_t = \sum_{k=1}^{K} \textbf{a}_{k,t}^{\intercal} \tilde{\textbf{X}} \textbf{w}_{k}^{out},
    \label{eq:gatsm}
\end{equation}
where $K$ is the number of attention heads, $\textbf{a}_{k,t} = \left[ a_{k,i,1}, a_{k,i,2}, \cdots, a_{k,i,T} \right] \in \mathbb{R}^{T}$ is the attention map defined in Equation~(\ref{eq:mha}), $\tilde{\textbf{X}} = \left[ \tilde{\textbf{x}}_1, \tilde{\textbf{x}}_2, \cdots, \tilde{\textbf{x}}_T \right] \in \mathbb{R}^{T \times M}$ are the transformed features defined in Equation~(\ref{eq:nbm}), and $\textbf{w}_{k}^{out} \in \mathbb{R}^{M}$ is the learnable output weight.

\subsection{Interpretability}
We can rewrite Equation~(\ref{eq:gatsm}) as the following scalar form:
\begin{equation}
    \begin{split}
        \sum_{k=1}^{K}& \textbf{a}_{k,t}^{\intercal} \tilde{\textbf{X}} \textbf{w}_{k}^{out}\\ &=
        \sum_{u=1}^{t} \sum_{m=1}^{M} \sum_{k=1}^{K} \sum_{b=1}^{B} a_{k,t,u} h_b \left( x_{u,m} \right) w_{m,b}^{nbm} w_{k,m}^{out} \\
        &= \sum_{u=1}^{t} \sum_{m=1}^{M} f_{u,m} \left( x_{u,m}, \textbf{X}_{:t} \right)
    \end{split}
    \label{eq:gatsm2}
\end{equation}
Equation~(\ref{eq:gatsm2}) shows that GATSM satisfies Definition 3.1 by encapsulating the attention scores (i.e., $\alpha_{k,t,u}$) and time-sharing NBM (i.e., $h_b$) into a function $f_{u,m}$. We can derive three types of interpretations from GATSM: 1) $a_{k,t,u}$ indicates the importance of time step $u$ at time step $t$, 2) $h_{b} \left( x_{u,m} \right) w_{m,b}^{nbm} w_{k,m}^{out}$ represents the time-independent contribution of feature $m$, and 3) $a_{k,t,u} h_{b} \left( x_{u,m} \right) w_{m,b}^{nbm} w_{k,m}^{out}$ represents the time-dependent contribution of feature $m$ at time step $t$.

\section{Experiments}

\subsection{Datasets}
We conducted experiments using eight publicly available real-world time series datasets. From the Monash repository \cite{tan_monash_2020}, we sourced three datasets: Energy, Rainfall, and AirQuality. We downloaded another three datasets, Heartbeat, LSST, and NATOPS, from the UCR repository \cite{bagnall_uea_2018}. We downloaded the remaining two datasets, Mortality and Sepsis, from the PhysioNet \cite{goldberger_physiobank_2000}. We performed ordinal encoding for categorical features and standardize features to have zero-mean and unit-variance. For regression task, we also standardized the target value $y$ to have a zero-mean and unit-variance. If the dataset contains missing values, we impute categorical features with their modes and numerical features with their means. We split the dataset into a 60\%/20\%/20\% ratio for training, validation, and testing, respectively.
Appendix~\ref{sec:dataset} provides further details about the experimental datasets, including their statistics and descriptions.

\subsection{Baselines}
We compare our GATSM with 15 baselines, which can be categorized into five groups: 1) Black-box tabular models including extreme gradient boosting (XGBoost) \cite{chen_xgboost_2016} and MLP; 2) Black-box time series models including simple recurrent neural network (RNN), gated recurrent unit (GRU), long short-term memory (LSTM), and Transformer \cite{ashish_vaswani_attention_2017}; 3) Transparent tabular models including simple linear model (Linear), explainable boosting machine (EBM) \cite{nori_interpretml_2019}, NAM \cite{rishabh_agarwal_neural_2021}, NodeGAM \cite{chun-hao_chang_node-gam_2022}, and NBM \cite{filip_radenovic_neural_2022}; 4) A transparent time series model, NATM \cite{jo_neural_2023}; 5) Multi-step forecasting specialized models including DLinear \cite{zeng_transformers_2023}, PatchTST \cite{nie_time_2023}, and iTransformer \cite{yong_itransformer_2024}.

\subsection{Implementation}
We implemented XGBoost and EBM models using the \texttt{xgboost} \cite{chen_xgboost_2016} and \texttt{interpretml} \cite{nori_interpretml_2019} libraries, respectively. For NodeGAM, we employ the official implementation provided by its authors \cite{chun-hao_chang_node-gam_2022}. We developed the remaining models using \texttt{torch} \cite{paszke_pytorch_2019}. All models undergo hyperparameter tuning via \texttt{optuna} \cite{akiba_optuna_2019}. The pytorch-based models are optimized with the Adam with decoupled weight decay (AdamW) \cite{loshchilov_decoupled_2019} optimizer on an NVIDIA A100 GPU. Model training is halted if the validation loss does not decrease over 20 epochs. We used mean squared error for the regression task, binary cross-entropy for the binary classification task, and cross-entropy for the multi-class classification task. Appendix~\ref{sec:implementation} provides further details of the model implementations and hyper-parameters.

\subsection{Comparison with baselines}\label{sec:performance}

\begin{table*}[!t]
    
    \centering

    \resizebox{.95\linewidth}{!}{
    
        \begin{tabular}{c|c|cccccccc|c}

            \toprule

            Model Type &
            Model &
            \makecell{Energy\\(\text{$R^2$}$\uparrow$)} &
            \makecell{Rainfall\\(\text{$R^2$}$\uparrow$)} &
            \makecell{AirQuality\\(\text{$R^2$}$\uparrow$)} &
            \makecell{Heartbeat\\(AUROC$\uparrow$)} &
            \makecell{Mortality\\(AUROC$\uparrow$)} &
            \makecell{Sepsis\\(AUROC$\uparrow$)} &
            \makecell{LSST\\(Acc$\uparrow$)} &
            \makecell{NATOPS\\(Acc$\uparrow$)} &
            Avg. Rank \\

            \midrule
            
            \multirow{4}{*}{\makecell{Black-box\\Tabular Model}}
            
            & \multirow{2}{*}{XGBoost} & 0.094 & 0.002 & 0.532 & 0.679 & 0.707 & \textbf{0.816} & 0.424 & 0.200 & 9.75 \\
            & & ($\pm$0.137) & ($\pm$0.002) & ($\pm$0.019) & ($\pm$0.094) & ($\pm$0.015) & ($\pm$0.007) & ($\pm$0.012) & ($\pm$0.049) & ($\pm$4.773) \\
            
            \arrayrulecolor{black!30}\cmidrule{2-11}
            
            & \multirow{2}{*}{MLP}
            & 0.459 & 0.011 & 0.423 & 0.654 & 0.842 & 0.786 & 0.417 & 0.211 & 8.875 \\
            & & ($\pm$0.101) & ($\pm$0.004) & ($\pm$0.031) & ($\pm$0.082) & ($\pm$0.014) & ($\pm$0.007) & ($\pm$0.008) & ($\pm$0.065) & ($\pm$3.044) \\
            
            \arrayrulecolor{black!30}\midrule
            
            \multirow{8}{*}{\makecell{Black-box\\Time Series Model}} & \multirow{2}{*}{RNN}
            & 0.320 & 0.068 & 0.644 & 0.661 & 0.581 & 0.782 & 0.422 & 0.592 & 8.500 \\
            & & ($\pm$0.122) & ($\pm$0.020) & ($\pm$0.032) & ($\pm$0.078) & ($\pm$0.040) & ($\pm$0.009) & ($\pm$0.029) & ($\pm$0.110) & ($\pm$2.673) \\
            
            \arrayrulecolor{black!30}\cmidrule{2-11}
            
            & \multirow{2}{*}{GRU}
            & 0.435 & 0.089 & \underline{0.701} & 0.694 & 0.818 & 0.785 & \underline{0.629} & 0.931 & 4.500 \\
            & & ($\pm$0.107) & ($\pm$0.034) & ($\pm$0.018) & ($\pm$0.052) & ($\pm$0.014) & ($\pm$0.010) & ($\pm$0.013) & ($\pm$0.045) & ($\pm$2.619) \\
            
            \arrayrulecolor{black!30}\cmidrule{2-11}
            
            & \multirow{2}{*}{LSTM}
            & 0.359 & \underline{0.090} & 0.683 & 0.648 & 0.790 & 0.779 & 0.491 & 0.908 & 6.875 \\
            & & ($\pm$0.112) & ($\pm$0.031) & ($\pm$0.026) & ($\pm$0.042) & ($\pm$0.020) & ($\pm$0.008) & ($\pm$0.082) & ($\pm$0.035) & ($\pm$3.603) \\
            
            \arrayrulecolor{black!30}\cmidrule{2-11}
            
            & \multirow{2}{*}{Transformer}
            & 0.263 & \textbf{0.098} & \textbf{0.711} & 0.690 & 0.844 & 0.789 & \textbf{0.679} & \textbf{0.967} & \underline{4.125} \\
            & & ($\pm$0.263) & ($\pm$0.035) & ($\pm$0.027) & ($\pm$0.040) & ($\pm$0.019) & ($\pm$0.010) & ($\pm$0.019) & ($\pm$0.029) & ($\pm$3.980) \\

            \arrayrulecolor{black!30}\midrule

            \multirow{6}{*}{\makecell{Forecasting Model}}
            & \multirow{2}{*}{DLinear}
            & 0.058 & 0.033 & 0.473 & 0.672 & \multirow{2}{*}{N/A} & \multirow{2}{*}{N/A} & 0.315 & 0.933 & 9.833\\
            && ($\pm$0.247) & ($\pm$0.012) & ($\pm$0.023) & ($\pm$0.074) & & & ($\pm$0.010) & ($\pm$0.018) & ($\pm$4.355)\\

            \arrayrulecolor{black!30}\cmidrule{2-11}

            & \multirow{2}{*}{PatchTST}
            & 0.312 & 0.056 & 0.488 & 0.578 & \multirow{2}{*}{N/A} & \multirow{2}{*}{N/A} & 0.589 & 0.769 & 8.666\\
            & & ($\pm$0.172) & ($\pm$0.016) & ($\pm$0.032) & ($\pm$0.082) & & & ($\pm$0.027) & ($\pm$0.058) & ($\pm$4.367)\\

            \arrayrulecolor{black!30}\cmidrule{2-11}

            & \multirow{2}{*}{iTransformer}
            & -0.080 & 0.055 & 0.627 & 0.642 & \multirow{2}{*}{N/A} & \multirow{2}{*}{N/A} & 0.538 & 0.797 & 8.833\\
            & & ($\pm$0.309) & ($\pm$0.012) & ($\pm$0.049) & ($\pm$0.047) & & & ($\pm$0.040) & ($\pm$0.041) & ($\pm$4.491)\\

            \arrayrulecolor{black!30}\midrule
            
            \multirow{10}{*}{\makecell{Transparent\\Tabular Model}}
            & \multirow{2}{*}{Linear}
            & \underline{0.482} & 0.004 & 0.241 & 0.637 & 0.838 & 0.723 & 0.311 & 0.206 & 11.875 \\
            & & ($\pm$0.112) & ($\pm$0.001) & ($\pm$0.019) & ($\pm$0.070) & ($\pm$0.017) & ($\pm$0.011) & ($\pm$0.010) & ($\pm$0.045) & ($\pm$4.941) \\
            
            \arrayrulecolor{black!30}\cmidrule{2-11}
            
            & \multirow{2}{*}{EBM}
            & -0.200 & 0.004 & 0.324 & 0.666 & 0.729 & 0.802 & 0.408 & 0.164 & 11.625 \\
            & & ($\pm$0.409) & ($\pm$0.001) & ($\pm$0.014) & ($\pm$0.056) & ($\pm$0.017) & ($\pm$0.011) & ($\pm$0.016) & ($\pm$0.053) & ($\pm$4.406) \\
            
            \arrayrulecolor{black!30}\cmidrule{2-11}
            
            & \multirow{2}{*}{NAM}
            & 0.363 & 0.006 & 0.300 & 0.645 & \underline{0.853} & 0.800 & 0.400 & 0.242 & 9.375\\
            && ($\pm$0.218) & ($\pm$0.002) & ($\pm$0.013) & ($\pm$0.026) & ($\pm$0.014) & ($\pm$0.006) & ($\pm$0.011) & ($\pm$0.040) & ($\pm$4.719) \\
            
            \arrayrulecolor{black!30}\cmidrule{2-11}
            
            & \multirow{2}{*}{NodeGAM}
            & 0.398 & 0.006 & 0.380 & 0.681 & \textbf{0.854} & \underline{0.802} & 0.400 & 0.247 & 7.750 \\
            & & ($\pm$0.195) & ($\pm$0.002) & ($\pm$0.032) & ($\pm$0.046) & ($\pm$0.013) & ($\pm$0.007) & ($\pm$0.028) & ($\pm$0.012) & ($\pm$4.892) \\
            
            \arrayrulecolor{black!30}\cmidrule{2-11}
            
            & \multirow{2}{*}{NBM}
            & 0.330 & 0.007 & 0.301 & 0.716 & 0.852 & 0.799 & 0.388 & 0.189 & 9.250 \\
            & & ($\pm$0.251) & ($\pm$0.003) & ($\pm$0.012) & ($\pm$0.039) & ($\pm$0.014) & ($\pm$0.006) & ($\pm$0.014) & ($\pm$0.029) & ($\pm$4.892) \\
            
            \arrayrulecolor{black!30}\midrule
            
            \multirow{4}{*}{\makecell{Transparent\\Time Series Model}}
            & \multirow{2}{*}{NATM}
            & 0.304 & 0.038 & 0.548 & \underline{0.724} & \multirow{2}{*}{N/A} & \multirow{2}{*}{N/A} & 0.452 & 0.878 & 6.833 \\
            & & ($\pm$0.122) & ($\pm$0.011) & ($\pm$0.028) & ($\pm$0.043) &  &  & ($\pm$0.010) & ($\pm$0.058) & ($\pm$2.927) \\
            
            \arrayrulecolor{black!30}\cmidrule{2-11}
            
            & \multirow{2}{*}{GATSM (ours)}
            & \textbf{0.493} & 0.073 & 0.583 & \textbf{0.843} & \underline{0.853} & 0.797 & 0.570 & \underline{0.956} & \textbf{3.375} \\
            & & ($\pm$0.173) & ($\pm$0.027) & ($\pm$0.026) & ($\pm$0.025) & ($\pm$0.015) & ($\pm$0.007) & ($\pm$0.024) & ($\pm$0.027) & ($\pm$1.996) \\
            
            \arrayrulecolor{black}\bottomrule
             
        \end{tabular}
    
    }

    \caption{Predictive performance comparison of various models on single output tasks.}
    
    \label{tab:performance}
    
\end{table*}

Table~\ref{tab:performance} shows the predictive performances of the experimental models. We report mean scores and standard deviations over five different random seeds. For the regression datasets, we evaluate $R^2$ scores. For the binary classification datasets, we assess the area under the receiver operating characteristic curve (AUROC). For the multi-class classification datasets, we measure accuracy. We highlight the best-performing model in \textbf{bold} and \underline{underlined} the second-best model. Since the tabular models cannot handle time series, they only take $\textbf{x}_T$ to produce $y_T$.

On the Energy and Heartbeat datasets, which have small sample sizes, our GATSM demonstrates the best performance, indicating strong generalization ability. EBM, XGBoost, and Transformer struggle with overfitting on the Energy dataset. For the Mortality and Sepsis datasets, there is no significant performance difference between tabular and time series models, nor between black-box and transparent models. This suggests that these two datasets lack significant temporal patterns and feature interactions. It is likely that seasonal patterns are hard to detect in medical data, and the patients' current condition already encapsulates previous conditions, making historical data less crucial. Since these datasets contain variable-length time series, the performances of NATM, which can only handle fixed-length time series, is not available. On the Rainfall, AirQuality, LSST, and NATOPS datasets, the time series models significantly outperform the tabular models, indicating that these datasets contain important temporal patterns that tabular models cannot capture. Additionally, the black-box models outperform the transparent models, suggesting that these datasets have higher-order feature interactions that transparent models cannot capture. Nevertheless, GATSM is the best model within the transparent model group and performs comparably to Transformer. This result implies that GATSM effectively captures the effects of input features and temporal patterns while maintaining transparency. Overall, GATSM achieved the best average rank in the experiments, followed by the Transformer, highlighting GATSM's robustness.

We also validated three state-of-the-art forecasting models, DLinear, PatchTST, and iTransformer. Experimental results indicate that these forecasting models do not consistently outperform the other comparative models. This is likely because their architectures are primarily optimized for forecasting tasks rather than regression or classification, and their large model sizes make them susceptible to overfitting. The performances for the three forecasting models on the Mortality and Sepsis datasets are unavailable because these models cannot handle variable-length many-to-many prediction tasks as same as NATM. We provide additional experiments on multi-step forecasting and synthetic tumor size datasets in Appendix~\ref{sec:additional_exp}.

\subsection{Ablation study}

\subsubsection{Choice of feature function}
We evaluate the performance of GATSM by changing the feature functions to three models: Linear, NAM, and NBM. Table~\ref{tab:ablation1} presents the results of this experiment. The simple linear function performs poorly because it lacks the capability to capture non-linear relationships. In contrast, NAM, which can capture non-linearity, shows improved performance over the linear function. However, NBM stands out by achieving the best performance in six out of eight datasets. This indicates that the basis strategy of NBM is highly effective for time series data.

\subsubsection{Design of temporal module}
We evaluate the performance of GATSM by modifying the design of the temporal module. Table~\ref{tab:ablation2} presents the results. GATSM without the temporal module (Base) fails to learn temporal patterns and shows poor performance in the experiment. GATSM with only positional encoding (Base + PE) also shows performance similar to the Base, indicating that positional encoding alone is insufficient for capturing effective temporal patterns. GATSM with only MHA (Base + MHA) outperforms the previous two methods, demonstrating that the MHA is beneficial for capturing temporal patterns. Finally, our full GATSM (Base + PE + MHA) significantly outperforms the other methods, suggesting that the combination of PE and MHA creates a synergistic effect. Consistent with our previous findings in Section~\ref{sec:performance}, all four methods show similar performances on the Mortality and Sepsis datasets, which lack significant temporal patterns.

\begin{table*}[!t]
    
    \centering

    \resizebox{\linewidth}{!}{
    
        \begin{tabular}{c|cccccccc}

            \toprule

            Feature Function & Energy & Rainfall & AirQuality & Heartbeat & Mortality & Sepsis & LSST & NATOPS \\

            \midrule

            Linear
            & 0.283($\pm$0.277) & 0.071($\pm$0.024) & 0.563($\pm$0.019) & 0.766($\pm$0.024) & 0.832($\pm$0.015) & 0.735($\pm$0.012) & 0.398($\pm$0.030) & \textbf{0.972}($\pm$0.020) \\

            NAM
            & 0.304($\pm$0.229) & 0.068($\pm$0.021) & 0.564($\pm$0.019) & 0.838($\pm$0.032) & 0.851($\pm$0.013) & \textbf{0.801}($\pm$0.005) & 0.553($\pm$0.023) & 0.933($\pm$0.039) \\

            NBM
            & \textbf{0.493}($\pm$0.173) & \textbf{0.073}($\pm$0.027) & \textbf{0.583}($\pm$0.026) & \textbf{0.843}($\pm$0.025) & \textbf{0.853}($\pm$0.015) & 0.797($\pm$0.007) & \textbf{0.570}($\pm$0.024) & 0.956($\pm$0.027) \\

            \bottomrule
            
        \end{tabular}

    }

    \caption{Ablation study on different feature functions.}
    
    \label{tab:ablation1}
    
\end{table*}
\begin{table*}[!t]
    
    \centering

    \resizebox{\linewidth}{!}{
    
        \begin{tabular}{c|cccccccc}

            \toprule

            Temporal Module & Energy & Rainfall & AirQuality & Heartbeat & Mortality & Sepsis & LSST & NATOPS \\

            \midrule

            Base & 0.452($\pm$0.087) & 0.007($\pm$0.002) & 0.299($\pm$0.012) & 0.661($\pm$0.043) & \textbf{0.854}($\pm$0.013) & 0.798($\pm$0.008) & 0.392($\pm$0.006) & 0.192($\pm$0.027) \\

            Base + PE & 0.397($\pm$0.054) & 0.007($\pm$0.003) & 0.299($\pm$0.012) & 0.681($\pm$0.068) & 0.852($\pm$0.013) & \textbf{0.799}($\pm$0.007) & 0.385($\pm$0.027) & 0.228($\pm$0.029) \\

            Base + MHA & 0.368($\pm$0.230) & 0.048($\pm$0.017) & 0.555($\pm$0.020) & 0.821($\pm$0.044) & 0.847($\pm$0.020) & 0.779($\pm$0.033) & \textbf{0.595}($\pm$0.013) & 0.856($\pm$0.059) \\

            Base + PE + MHA & \textbf{0.493}($\pm$0.173) & \textbf{0.073}($\pm$0.027) & \textbf{0.583}($\pm$0.026) & \textbf{0.843}($\pm$0.025) & 0.853($\pm$0.015) & 0.797($\pm$0.007) & 0.570($\pm$0.024) & \textbf{0.956}($\pm$0.027) \\

            \bottomrule
            
        \end{tabular}

    }

    \caption{Ablation study on the temporal module.}
    
    \label{tab:ablation2}
    
\end{table*}
\begin{table*}[!t]
    
    \centering
    
    \resizebox{\linewidth}{!}{
    
        \begin{tabular}{c|cccccccc}
            \toprule
            & Energy & Rainfall & AirQuality & Heartbeat & Mortality & Sepsis & LSST & NATOPS\\
            \midrule
            Learnable PE & 0.519($\pm$0.097) & 0.072($\pm$0.026) & 0.570($\pm$0.023) & 0.838($\pm$0.048) & 0.855($\pm$0.018) & 0.797($\pm$0.006) & 0.570($\pm$0.023) & 0.950($\pm$0.021) \\
            Sinusoidal PE & 0.493($\pm$0.173) & 0.073($\pm$0.027) & 0.583($\pm$0.026) & 0.843($\pm$0.025) & 0.853($\pm$0.015) & 0.797($\pm$0.007) & 0.570($\pm$0.024) & 0.956($\pm$0.027) \\
            \bottomrule
        \end{tabular}
    
    }

    \caption{Performances of learnable position embedding and sinusoidal positional encoding on GATSM.}

    \label{tab:ablation3}
        
\end{table*}

\subsubsection{Positional Encoding}
As described in Section~\ref{sec:mha}, we recommend sinusoidal positional encoding over learnable position embeddings because the maximum length of time series is often unknown. In contrast, learnable position embeddings require knowledge of the maximum length, which is typically infeasible in practice. Nevertheless, we conducted an experiment comparing sinusoidal positional encoding and learnable position embedding to determine if significant performance differences exist. Table~\ref{tab:ablation3} presents the performance comparison between learnable position embedding and sinusoidal positional encoding on GATSM. The experimental result demonstrating no significant difference between these two methods.

\begin{table*}[!t]
    
    \centering

    \resizebox{.7\linewidth}{!}{
    
        \begin{tabular}{c|cccccccc}

            \toprule
        
             & Energy & Rainfall & AirQuality & Heartbeat & Mortality & Sepsis & LSST & NATOPS \\

             \midrule
             
             NAM & 65.3K & 1.8M & 5.1M & 139.1K & 772.2K & 23.9K & 2.3M & 147.9K \\
             NBM & 45.5K & 1.1M & 1.0M & 55.9K & 375.8K & 6.5K & 1.6M & 85.6K \\
             Transformer & 30.9K & 240.5K & 174.2K & 15.7K & 161.9K & 134.6K & 214.4K & 68.3K \\
             NATM & 5.3K & 699.3K & 241.3K & 1.3K & N/A & N/A & 28.6K & 19.2K \\
             GATSM & 6.1K & 350.6K & 192.8K & 1.2K & 4.9K & 3.8K & 126.5K & 12.5K \\

             \bottomrule
             
        \end{tabular}

    }

    \caption{Inference throughputs per second of different models.}
    
    \label{tab:throughput}
    
\end{table*}

\begin{table*}[!t]
    
    \centering

    \resizebox{.7\linewidth}{!}{
    
        \begin{tabular}{c|cccccccc}

            \toprule
        
             & Energy & Rainfall & AirQuality & Heartbeat & Mortality & Sepsis & LSST & NATOPS \\

             \midrule
             
             NAM & 307.2K & 39.36K & 115.2K & 780.8K & 524.8K & 172.03M & 81.79K & 314.88K\\
             NBM & 5.39M & 677.65K & 2.03M & 13.78M & 9.26M & 3.04G & 1.35M & 5.39M\\
             Transformer & 5.3M & 5.14M & 19.4M & 32.72M & 1.29M & 3.79M & 1.32M & 8M\\
             NATM & 44.25M & 921.89K & 2.77M & 316.32M & N/A & N/A & 2.84M & 15.68M\\
             GATSM & 777.19M & 16.2M & 48.58M & 5.59G & 1.87G & 3.04G & 48.62M & 276.84M\\

             \bottomrule
             
        \end{tabular}

    }

    \caption{FLOPs of different models.}
    
    \label{tab:flops}
    
\end{table*}

\subsection{Inference speed}

The inference speed of machine learning models is a crucial metric for real-world systems. We evaluate the throughput and FLOPs of the models. Table~\ref{tab:throughput} and Table~\ref{tab:flops} present the results. Since all experimental datasets have fewer features than the number of basis functions in NBM, NAM achieves higher throughput than NBM. Transparent tabular models typically exhibit fast speeds. However, their throughput significantly decreases in datasets with many features, such as Heartbeat, Mortality, and Sepsis, because they require the same number of feature functions as the number of input features. Transformer shows higher throughput than the transparent time series models because it does not require feature functions, which are the main bottleneck of transparent models. Additionally, the PyTorch implementation of Transformer uses the flash attention mechanism \cite{dao_flashattention_2022} to enhance its efficiency. NATM has slightly higher throughput than GATSM, as it does not require the attention mechanism and has fewer feature functions compared to the number of basis functions in GATSM.

Inference latency requirements in real-world applications vary significantly based on the context. For instance, real-time clinical decision support systems, such as heart failure prediction models, typically demand latencies below one second per patient, whereas non-real-time applications, such as discharge prediction, have less stringent requirements. As demonstrated in Table~\ref{tab:throughput}, GATSM can process at least 1,000 samples per second, making it feasible for real-world scenarios.

\begin{figure*}[!t]
    \centering
    \begin{subfigure}{.32\linewidth}
        \centering
        \includegraphics[width=\linewidth]{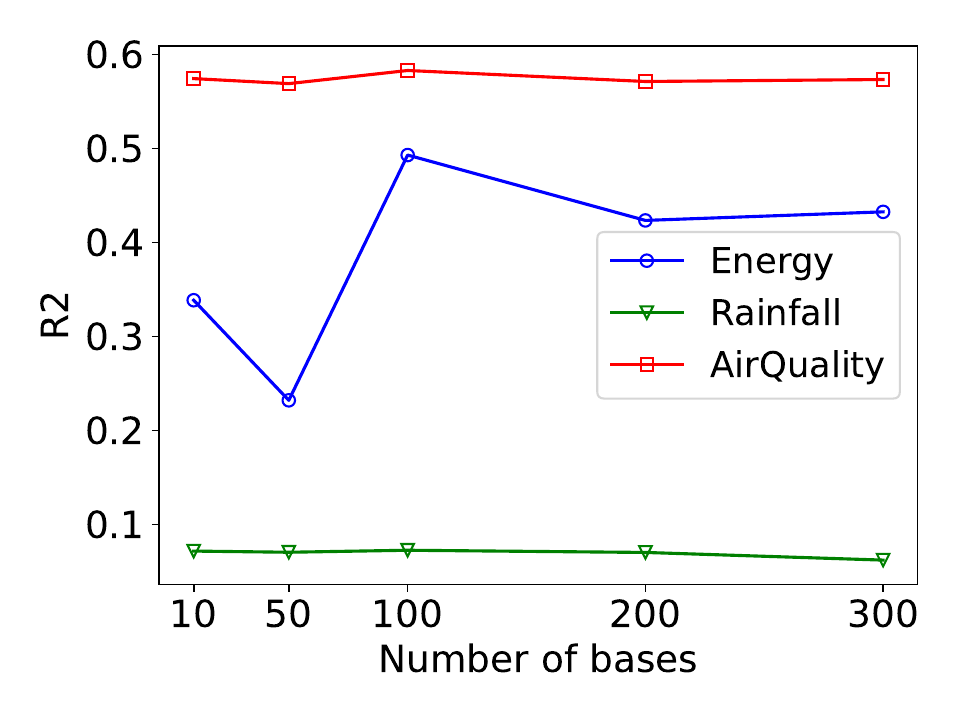}
        \caption{Forecasting}
    \end{subfigure}
    \begin{subfigure}{.32\linewidth}
        \centering
        \includegraphics[width=\linewidth]{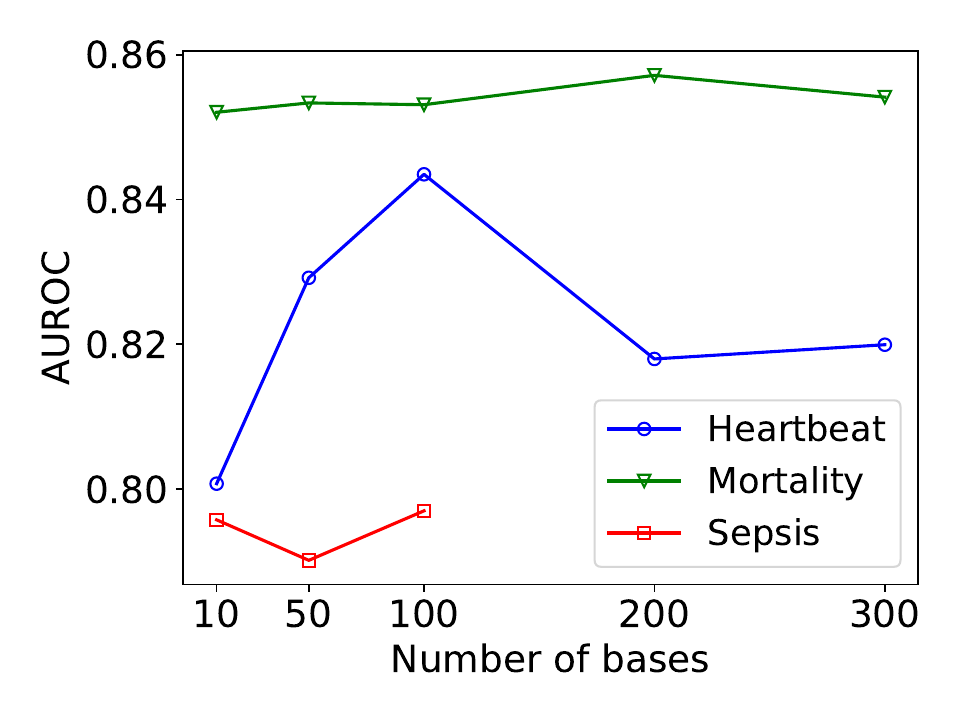}
        \caption{Binary}
    \end{subfigure}
    \begin{subfigure}{.32\linewidth}
        \centering
        \includegraphics[width=\linewidth]{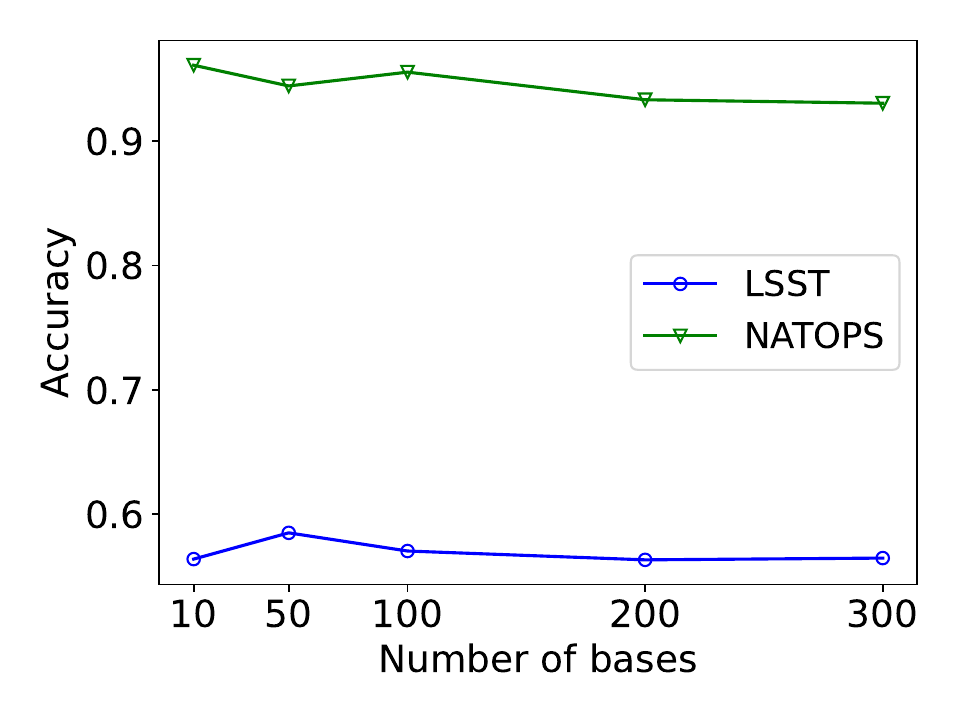}
        \caption{Multi-class}
    \end{subfigure}
    \caption{Performances of GATSM on the different number of basis functions.}
    \label{fig:basis}
\end{figure*}

\subsection{Number of basis functions}
We evaluate GATSM by varying the number of basis functions in the time-sharing NBM. Figure~\ref{fig:basis} presents the results for regression, binary classification, and multi-class classification datasets. For the Sepsis dataset, using 200 and 300 basis functions causes an out-of-memory error. For the Energy and Heartbeat datasets, performance improves up to 100 basis functions but shows no further benefit when the number of basis functions exceeds 100. In other datasets, performance changes are not significant with different numbers of basis functions. In addition, there is a trade-off between the number of basis functions and computational speed. Therefore, we recommend generally setting the number of basis functions to 100. Note that the performance of GATSM with this hyper-parameter depends on the dataset size and complexity. Hence, a larger number of basis functions may be beneficial for more complex datasets.

\subsection{Interpretation}\label{sec:interpretation}
In this section, we describe four interpretations of GATSM's predictions on the AirQuality dataset. Due to page limits, visualizations for model interpretation are provided in Appendix~\ref{sec:vis}. In addition, interpretations for the Rainfall and Mortality datasets can be found in Appendix~\ref{sec:vis}.

\subsubsection{Time-step importance}
We plot the average attention scores at the last time step $T$ in Figure~5. The process for extracting the average attention score of time step $u$ at time step $t$ is formalized as $\sum_{k=1}^{K} a_{k,t,u}$. This process is repeated over all data samples, and the results are averaged. Based on Figure~5, it seems that GATSM pays more attention to the initial and last states than to the intermediate states. This indicates that the current concentration of particulate matter depends on the initial and recent states.

\subsubsection{Global feature contribution}
Figure~6 illustrates the global behavior of features in the AirQuality dataset, with red bars indicating the density of training samples. We extract $\sum_{k=1}^{K} h_b \left( x_{t,m} \right) w_{m,b}^{nbm} w_{k,m}^{out}$ from GATSM and repeat this process over the range of minimum to maximum feature values to plot the line. We found that the behavior of \textit{SO2}, \textit{O3}, and \textit{windspeed} is inconsistent with prior human knowledge. Typically, high levels of \textit{SO2} and \textit{O3} are associated with poor air quality. However, GATSM learned that particulate matter concentration starts to decrease when \textit{SO2} exceeds 10 and \textit{O3} exceeds 5. This discrepancy may be due to sparse training samples in these regions, leading to insufficient training, or there may be interactions with other features. Another known fact is that high \textit{windspeed} decreases particulate matter concentration. This is consistent when \textit{windspeed} is below 0.7 in our observation. However, particulate matter concentration drastically increases when \textit{windspeed} exceeds 0.7, likely due to the wind causing yellow dust.

\subsubsection{Local time-independent feature contribution}
To interpret the prediction of a data sample, we plot the local time-independent feature contributions, $\sum_{k=1}^{K} h_b \left( x_{t,m} \right) w_{m,b}^{nbm} w_{k,m}^{out}$, in Figure~7. The main y-axis (\textcolor{blue}{blue}) represents feature contribution, the sub y-axis (\textcolor{red}{red}) represents feature value, and the x-axis represents time steps. We found that \textit{SO2}, \textit{NO2}, \textit{CO}, and \textit{O3} have positive correlations. In contrast, \textit{temperature}, \textit{pressure}, \textit{dew point}, and \textit{windspeed} have negative correlations. These are consistent with the global interpretations shown in Figure~6. Rainfall has the same values across all time steps.

\subsubsection{Local time-dependent feature contribution}
We also visualize the local time-dependent feature contributions, $\sum_{k=1}^{K} a_{k,t,u} h_b \left( x_{t,m} \right) w_{m,b}^{nbm} w_{k,m}^{out}$. Figure~8 illustrates the interpretation of the same data sample as in Figure~7. The time-dependent interpretation differs slightly from the time-independent interpretation. We found that there are time lags in \textit{SO2}, \textit{NO2}, \textit{CO}, and \textit{O3}, meaning previous feature values affect current feature contributions. For example, in the case of \textit{SO2}, low feature values around time step 5 lead to low feature contributions around time step 13.

\section{Conclusion}
In this paper, we proposed a novel transparent model for time series named GATSM. GATSM consists of time-sharing NBM and the temporal module to effectively learn feature representations and temporal patterns while maintaining transparency. The experimental results demonstrated that GATSM has superior generalization ability and is the only transparent model with performance comparable to Transformer. We provided various visual interpretations of GATSM, demonstrated that GATSM capture interesting patterns in time series data. GATSM delivers robust predictive performance along with detailed interpretations of its outputs. This advantage enables accurate and trustworthy predictions, which is crucial in high-stakes domains. We anticipate that GATSM will be widely adopted in various fields and demonstrate strong performance. We discuss the limitations of GATSM and suggest directions for future work in Appendix~\ref{sec:limitations}. Potential applications and the broader impacts of GATSM across various fields can be found in Appendix~\ref{sec:broader}.

\appendix

\section{Limitations \& Future Works}\label{sec:limitations}
Although GATSM achieved state-of-the-art performance within the transparent model category, it has several limitations. This section discusses these limitations and suggests future work to address them. GAMs have relatively slower computational times and larger model sizes compared to black-box models because they require the same number of feature functions as input features. To address this problem, methods such as the basis strategy can be proposed to reduce the number of feature functions, or entirely new methods for transparent models can be developed. The attention mechanism in GATSM may be a bottleneck. Fast attention mechanisms proposed in the literature \cite{katharopoulos_transformers_2020,madaan_treeformer_2023,sinong_wang_linformer_2020,krzysztof_choromanski_rethinking_2021,nikita_kitaev_reformer_2020,tri_flash_2024}, or the recently proposed Mamba \cite{gu_mamba_2023}, can help overcome this limitation. Existing time series models, including GATSM, only handle discrete time series and have limited length generalization ability, resulting in significantly reduced performance when very long sequences, unseen during training, are input. Extending GATSM to continuous models using NeuralODE \cite{chen_neural_2018} or HiPPO \cite{gu_hippo_2020} could address this issue. GATSM still cannot learn higher-order feature interactions internally and shows low performance on complex datasets. Feature interaction methods proposed for transparent models may help address this problem \cite{kim_higher-order_2022,dubey_scalable_2022}.

\section{Broader impact}\label{sec:broader}
We discuss the expected impacts of GATSM across various fields.
\begin{itemize}
    \item \textbf{Time series adaptation:} GATSM extends existing GAMs to time series, enabling tasks that traditional GAMs could not perform in this context - e.g., better performance on time series and finding temporal patterns.
    \item \textbf{Improved decision-making system:} GATSM can show users their exact decision-making process, providing trust and confidence in its predictions to users. This enables decision-makers to make more informed choices, crucial in high-stakes domains such as healthcare.
    \item \textbf{Ethical AI:} GATSM can examine that their outcomes are biased or discriminatory by displaying the shape of feature functions. This is particularly important in ethically sensitive domains, such as recidivism prediction.
    \item \textbf{Scientific discovery:} Researchers have already used transparent models in various research fields for scientific discovery \cite{pedersen_hierarchical_2019,hastie_generalized_1995}. GATSM also can be applied to these domains to obtain novel scientific insights.
\end{itemize}
Despite these advantages, it is important to remember that the interpretations of transparent models do not necessarily reflect exact causal relationships. While transparent models provide clear and faithful interpretations, they are still not capable of identifying causal relationships. Causal discovery is a complex task that requires further research.

\begin{table*}[!t]
        
    \centering

    \resizebox{.7\linewidth}{!}{
    
        \begin{tabular}{ccccccc}

            \toprule
        
            Dataset & Task & Variable length & \# of time series & Avg. length & \# of features & \# of classes \\

            \midrule
            
            Energy & Regression & No & 137 & 24 & 24 & - \\
            Rainfall & Regression & No & 160,267 & 24 & 3 & - \\
            AirQuality & Regression & No & 16,966 & 24 & 9 & - \\
            Heartbeat & Binary & No & 409 & 405 & 61 & 2 \\
            Mortality & Binary & Yes & 12,000 & 49.861 & 41 & 2 \\
            Sepsis & Binary & Yes & 40,336 & 38.482 & 40 & 2 \\
            LSST & Multi-class & No & 4,925 & 36 & 6 & 14 \\
            NATOPS & Multi-class & No & 360 & 51 & 24 & 6 \\
            \bottomrule
            
        \end{tabular}

    }

    \caption{Dataset statistics.}
    
    \label{tab:statistics}
    
\end{table*}

\section{Related Works}\label{sec:related_works}

Various XAI studies have been conducted over the past decade \cite{rishabh_agarwal_neural_2021,chun-hao_chang_node-gam_2022,filip_radenovic_neural_2022,sebastian_bach_pixel-wise_2015,shrikumar_learning_2017}. However, they are less relevant to the transparent model that this study focuses on. Therefore, we refer readers to \cite{ali_explainable_2023,hassija_interpreting_2024} for more detailed information on recent XAI research. In this section, we review existing transparent models closely related to our GATSM and discuss their limitations.
Originally, GAMs were fitted via the backfitting algorithm using smooth splines \cite{hastie_generalized_1986,wahba_spline_1990}. Later, \cite{yin_lou_intelligible_2012} and \cite{nori_interpretml_2019} proposed boosted decision tree-based GAMs, which use boosted decision trees as feature functions. Spline- and tree-based GAMs have less flexibility and scalability. Thus, extending them to transfer or multi-task learning is challenging. To overcome this problem, various neural network-based GAMs have been proposed in recent years. \cite{potts_generalized_1999} introduced generalized additive neural network, which employs 2-layer neural networks as feature functions. Similarly, \cite{rishabh_agarwal_neural_2021} proposed neural additive model (NAM) that employs multi-layer neural networks. To improve the scalability of NAM, \cite{chun-hao_chang_node-gam_2022} and \cite{filip_radenovic_neural_2022} proposed the neural oblivious tree-based GAM and the basis network-based GAM, respectively. \cite{xu_sparse_2023} introduced a sparse version of NAM using the group LASSO. One disadvantage of GAMs is their limited predictive power, which stems from the fact that they only learn first-order feature interactions -- i.e., relationships between the target value and individual features. To address this, various studies have been conducted to enhance the predictive powers of GAMs by incorporating higher-order feature interactions, while still maintaining transparency. GA$^2$M \cite{lou_accurate_2013} simply takes pairwise features as inputs to learn pairwise interactions. GAMI-Net \cite{zebin_yang_gami-net_2021}, a neural network-based GAM, consists of networks for main effects (i.e., first-order interactions) and pairwise interactions. To enhance the interpretability of GAMI-Net, the sparsity and heredity constraints are added, and trivial features are pruned in the training process. Sparse interaction additive network \cite{enouen_sparse_2022} is a 3-phase method for exploiting higher-order interactions. Initially, a black-box neural network is trained; subsequently, explainable feature attribution methods like LIME \cite{ribeiro_why_2016} and SHAP \cite{lundberg_unified_2017} identify the top-k important features, and finally, NAM is trained with these extracted features. \cite{dubey_scalable_2022} introduced scalable polynomial additive model, an end-to-end model that learns higher-order interactions via polynomials. Similarly, \cite{kim_higher-order_2022} proposed higher-order NAM that utilizes the feature crossing technique to capture higher-order interactions. Despite their capabilities, the aforementioned GAMs cannot process time series data, which limits their applicability in real-world scenarios. Recently, neural additive time series model (NATM) \cite{jo_neural_2023}, a time-series adaptation of NAM, has been proposed. However, NATM handles each time step independently with separate feature networks. This approach cannot capture temporal patterns and only takes a fixed-length time series as input. Our GATSM not only captures temporal patterns but also handles dynamic-length time series.

\section{Dataset details}\label{sec:dataset}

We use eight publicly available datasets for our experiments. Three datasets - Energy, Rainfall, and AirQuality - can be downloaded from the Monash repository \cite{tan_monash_2020}. Another three datasets - Heartbeat, LSST, and NATOPS - are available from the UCR repository \cite{bagnall_uea_2018}. The remaining two datasets can be downloaded from the PhysioNet \cite{goldberger_physiobank_2000}. Table~\ref{tab:statistics} shows the statistics of the experimental datasets. Details of the datasets are provided below:
\begin{itemize}
    \item \textbf{Energy} \cite{chang_wei_tan_2020_3902637}: This dataset consists of 24 features related to temperature and humidity from sensors and weather conditions. These features are measured every 10 minutes. The goal of this dataset is to predict total energy usage.
    \item \textbf{Rainfall} \cite{chang_wei_tan_2020_3902654}: This dataset consists of temperatures measured hourly. The goal of this dataset is to predict total daily rainfall in Australia.
    \item \textbf{AirQuality} \cite{chang_wei_tan_2020_3902667}: This dataset consists of features related to air pollutants and meteorological data. The goal of this dataset is to predict the PM10 level in Beijing.
    \item \textbf{Heartbeat} \cite{liu_classification_2016}: This dataset consists of heart sounds collected from various locations on the body. Each sound was truncated to five seconds, and a spectrogram of each instance was created with a window size of 0.061 seconds with a 70\% overlap. The goal of this dataset is to classify the sounds as either normal or abnormal.
    \item \textbf{Mortality} \cite{silva_predicting_2012} This dataset consists of records of adult patients admitted to the ICU. The input features include the patient demographics, vital signs, and lab results. The goal of this dataset is to predict the in-hospital death of patients.
    \item \textbf{Sepsis} \cite{reyna_early_2019}: This dataset consists of records of ICU patients. The input features include patient demographics, vital signs, and lab results. The goal of this dataset is to predict sepsis six hours in advance at every time step.
    \item \textbf{LSST} \cite{emille_plasticc_2018}: This challenge dataset aims to classify astronomical time series. These time series consist of six different light curves, simulated based on the data expected from the Large Synoptic Survey Telescope (LSST).
    \item \textbf{NATOPS} \cite{alonso_aaltd_2016}: This dataset aims to classify the Naval Air Training and Operating Procedures Standardization (NATOPS) motions used to control aircraft movements. It consists of 24 features representing the x, y, and z coordinates for each of the eight sensor locations attached to the body.
\end{itemize}
We used \texttt{get\_UCR\_data()} and \texttt{get\_Monash\_regression\_data()} functions in the \texttt{tsai=0.3.9} library \cite{tsai} to load the UCR and Monash datasets.

\section{Implementation details}\label{sec:implementation}

\begin{table*}[!t]
        
    \centering

    \resizebox{\linewidth}{!}{
    
        \begin{tabular}{c|cccccccc}

            \toprule

            \multicolumn{9}{l}{\textbf{GATSM:} [256, 256, 128] hidden dims, 100 basis functions} \\

            \midrule
            
            Dataset & Batch Size & NBM Batch Norm. & NBM Dropout & Attn. Embedding Size & Attn. Heads & Attn. Dropout & Learning Rate & Weight Decay \\
            
            \midrule
            
            Energy & 32 & False & 2.315e-1 & 110 & 8 & 6.924e-2 & 4.950e-3 & 1.679e-3 \\
            Rainfall & 32,768 & False & 5.936e-3 & 44 & 7 & 1.215e-3 & 9.225e-3 & 2.204e-6 \\
            AirQuality & 4,096 & False & 2.340e-2 & 81 & 8 & 1.169e-1 & 6.076e-3 & 5.047e-6 \\
            Heartbeat & 64 & True & 1.749e-1 & 92 & 2 & 1.653e-1 & 8.061e-3 & 4.787e-6 \\
            Mortality & 512 & False & 7.151e-2 & 125 & 8 & 7.324e-1 & 7.304e-3 & 2.181e-4 \\
            Sepsis & 512 & True & 6.523e-2 & 90 & 6 & 8.992e-1 & 4.509e-3 & 2.259e-2 \\
            LSST & 1,024 & False & 2.500e-2 & 59 & 7 & 2.063e-1 & 5.561e-2 & 5.957e-3 \\
            NATOPS & 64 & True & 4.827e-3 & 49 & 8 & 7.920e-1 & 8.156e-3 & 2.748e-2 \\
            
            \bottomrule
            
        \end{tabular}

    }

    \caption{Optimal hyper-parameters for GATSM.}
    
    \label{tab:hparam}
    
\end{table*}

We use 13 models, including GATSM, for our experiments. We implement XGBoost and EBM using the \texttt{xgboost=2.0.3} \cite{chen_xgboost_2016} and \texttt{interpretml=0.6.1} \cite{nori_interpretml_2019} libraries, respectively. For NodeGAM, we employ the official implementation provided by its authors \cite{chun-hao_chang_node-gam_2022}. We developed the remaining models using \texttt{torch=2.0.1} \cite{paszke_pytorch_2019}. In addition, we implement the feature functions in NAM and NBM using grouped convolutions \cite{krizhevsky_imagenet_2012,xie_aggregated_2017} to enhance their efficiency. We trained XGBoost and EBM on two AMD EPYC 7513 CPUs, while we trained the other models on an NVIDIA A100 GPU with 80GB VRAM. All models undergo hyperparameter tuning via \texttt{optuna=3.6.1} \cite{akiba_optuna_2019} with the Tree-structured Parzen Estimator (TPE) algorithm \cite{bergstra_algorithms_2011} in 100 trials. The hyperparameter search space and the optimal hyperparameters for the models are provided below:
\begin{itemize}
    \item \textbf{XGBoost:} We tune the \texttt{n\_estimators} in the integer interval [1, 1000], \texttt{max\_depth} in the integer interval [0, 2000], learning rate in the continuous interval [1e-6, 1], \texttt{subsample} in the continuous interval [0, 1], and \texttt{colsample\_bytree} in the continuous interval [0, 1].
    \item \textbf{MLP, NAM, NBM and NATM:} We tune the \texttt{batchnorm} in the descret set \{False, True\}, \texttt{dropout} in the continuous interval [0, 0.9], \texttt{learning\_rate} in the continuous interval [1e-3, 1e-2], and \texttt{weight\_decay} in the continuous interval [1e-6, 1e-1] on a log scale.
    \item \textbf{RNN, GRU and LSTM:} We tune the \texttt{hidden\_size} in the integer interval [8, 128], \texttt{dropout} in the continuous interval [0, 0.9], \texttt{learning\_rate} in the continuous interval [1e-3, 1e-2], and \texttt{weight\_decay} in the continuous interval [1e-6, 1e-1] on a log scale.
    \item \textbf{Transformer:} We tune the \texttt{n\_layers} in the integer interval [1, 4], \texttt{emb\_size} in the integer interval [8, 32], \texttt{hidden\_size} in the integer interval [8, 128], \texttt{n\_heads} in the integer interval [1, 8], \texttt{dropout} in the continuous interval [0, 0.9], \texttt{learning\_rate} in the continuous interval [1e-3, 1e-2], and \texttt{weight\_decay} in the continuous interval [1e-6, 1e-1] on a log scale.
    \item \textbf{Linear:} We tune the \texttt{learning\_rate} in the continuous interval [1e-3, 1e-2], and \texttt{weight\_decay} in the continuous interval [1e-6, 1e-1] on a log scale.
    \item \textbf{EBM:} We tune \texttt{max\_bins} in the integer interval [8, 512], \texttt{min\_samples\_leaf} and \texttt{max\_leaves} in the integer interval [1, 50], \texttt{inner\_bags} and \texttt{outer\_bags} in the integer interval [1, 128], \texttt{learning\_rate} in the continuous interval [1e-6, 100] on a log scale, and \texttt{max\_rounds} in the integer interval [1000, 10000].
    \item \textbf{NodeGAM:} We tune \texttt{n\_trees} in the integer interval [1, 256], \texttt{n\_layers} and \texttt{depth} in the integer intervals [1, 4], \texttt{dropout} in the continuous interval [0, 0.9], \texttt{learning\_rate} in the continuous interval [1e-3, 1e-2], and \texttt{weight\_decay} in the continuous interval [1e-6, 1e-1] on a log scale.
    \item \textbf{GATSM:} We tune \texttt{nbm\_batchnorm} in the descret set \{False, True\}, \texttt{nbm\_dropout} in the continuous interval [0, 0.9], \texttt{attn\_emb\_size} in the integer interval [8, 128], \texttt{attn\_n\_heads} in the integer interval [1, 8], \texttt{attn\_dropout} in the continuous interval [0, 0.9], \texttt{learning\_rate} in the continuous interval [1e-3, 1e-2], and \texttt{weight\_decay} in the continuous interval [1e-6, 1e-1] on a log scale. Table~\ref{tab:hparam} provides the optimal hyper-parameters for GATSM across all experimental datasets.
\end{itemize}

\section{Additional experiments}\label{sec:additional_exp}

\subsection{Multi-step forecasting}

\begin{table}[h]
    
    \centering

    \resizebox{.7\linewidth}{!}{
        \begin{tabular}{c|ccc}
            \toprule
             & Electricity & Traffic & Weather \\
             \midrule
             \# of features & 321 & 862 & 21 \\
             Length & 26,304 & 17,544 & 52,696 \\
             \bottomrule
        \end{tabular}
    }

    \caption{Statistics of the forecasting datasets.}
    
    \label{tab:forecasting_stats}
    
\end{table}
\begin{table*}[!t]
    
    \centering

    \resizebox{.8\linewidth}{!}{
    
        \begin{tabular}{c|ccc|ccc|ccc}
        
            \toprule
            
             \multirow{2}{*}{Model} & \multicolumn{3}{c|}{Electricity(MAPE$\downarrow$)} & \multicolumn{3}{c|}{Traffic(MAPE$\downarrow$)} & \multicolumn{3}{c}{Weather(MAPE$\downarrow$)} \\
             & 24h & 48h & 72h & 24h & 48h & 72h & 24h & 48h & 72h \\
             
             \midrule
             
             \multirow{2}{*}{DLinear} & 0.108 & 0.115 & 0.113 & 0.242 & 0.250 & 0.252 & 0.844 & 0.860 & 0.817 \\
             & ($\pm$0.003) & ($\pm$0.004) & ($\pm$0.001) & ($\pm$0.011) & ($\pm$0.007) & ($\pm$0.012) & ($\pm$0.085) & ($\pm$0.013) & ($\pm$0.032) \\
             
             \multirow{2}{*}{PatchTST} & 0.100 & 0.107 & 0.104 & 0.209 & 0.227 & 0.228 & 0.622 & 0.580 & 0.581 \\
             & ($\pm$0.005) & ($\pm$0.007) & ($\pm$0.005) & ($\pm$0.005) & ($\pm$0.013) & ($\pm$0.007) & ($\pm$0.006) & ($\pm$0.022) & ($\pm$0.019) \\
             
             \multirow{2}{*}{GATSM} & 0.122 & 0.135 & 0.137 & 0.314 & 0.302 & 0.347 & 0.859 & 0.966 & 0.935 \\
             & ($\pm$0.005) & ($\pm$0.010) & ($\pm$0.007) & ($\pm$0.030) & ($\pm$0.046) & ($\pm$0.108) & ($\pm$0.115) & ($\pm$0.182) & ($\pm$0.133) \\
             
             \bottomrule
             
        \end{tabular}
        
    }

    \caption{Predictive performance comparison on the forecasting datasets.}
    
    \label{tab:forecasting}
    
\end{table*}

Multi-step forecasting is a crucial task in time series. To validate the performance of GATSM on the multi-step forecasting task, we conducted experiments using two sophisticated forecasting models, DLinear\cite{zeng_transformers_2023} and PatchTST\cite{nie_time_2023}, on three datasets: Electricity\footnote{\url{https://archive.ics.uci.edu/ml/datasets/ElectricityLoadDiagrams20112014}}, Traffic\footnote{\url{http://pems.dot.ca.gov}}, and Weather\footnote{\url{https://www.bgc-jena.mpg.de/wetter/}}. Table~\ref{tab:forecasting_stats} provides statistics on the experimental datasets. The models use 672 hours (four weeks) of data as input to generate 24, 48, and 72 hours-ahead predictions.

Table~\ref{tab:forecasting} presents the mean average percentage errors (MAPEs) and standard deviations of the experimental models over five different random seeds. Overall, PatchTST demonstrates the highest predictive accuracy, while GATSM shows lower predictive performance compared to the two state-of-the-art black-box models, DLinear and PatchTST. However, our GATSM is a transparent model and offers a unique advantage in terms of interpretability.

\begin{table}[h]
    
    \centering
    
    \resizebox{.5\linewidth}{!}{
    
        \begin{tabular}{c|c}
        
            \toprule
            
            Model & Cancer(\text{$R^2$}$\uparrow$) \\
            
            \midrule
            
            DLinear & 0.956($\pm$0.009) \\
            PatchTST & 0.953($\pm$0.006) \\
            GATSM & 0.906($\pm$0.016) \\
            
            \bottomrule
            
        \end{tabular}
        
    }

    \caption{Predictive performance comparison on the synthetic Cancer dataset.}
    
    \label{tab:cancer}
    
\end{table}
\begin{figure*}[!t]
    \centering
    \begin{minipage}{.8\linewidth}
        \centering
        \includegraphics[width=.4\linewidth]{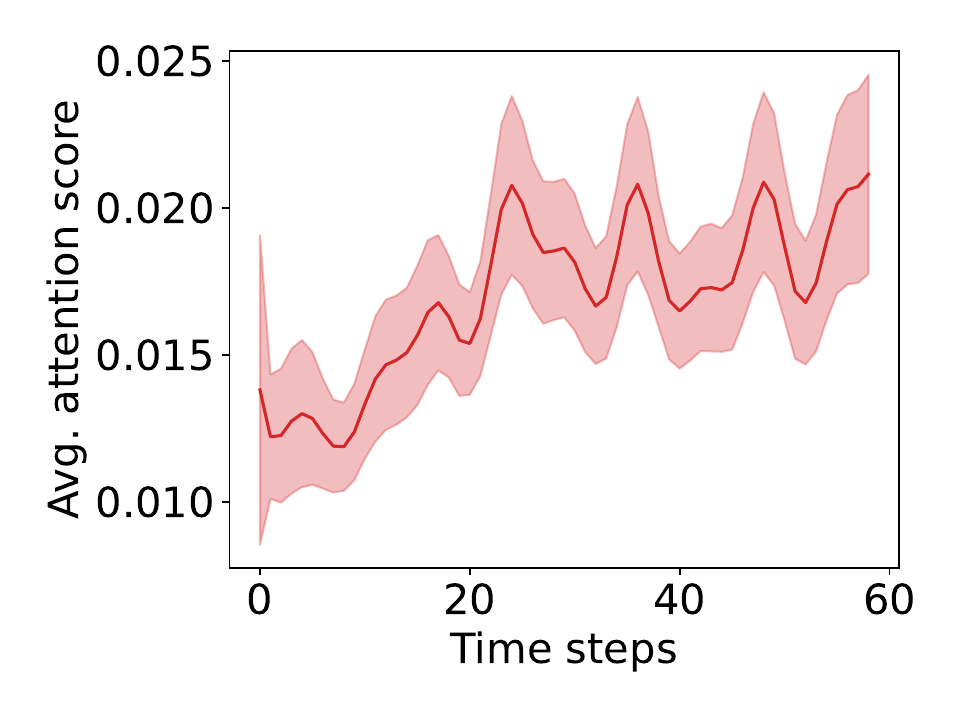}
        \caption{Average attention scores of time steps on the Cancer dataset.}
        \label{fig:attention_cancer}
    \end{minipage}
    \begin{minipage}{.75\linewidth}
        \centering
        \includegraphics[width=.9\linewidth]{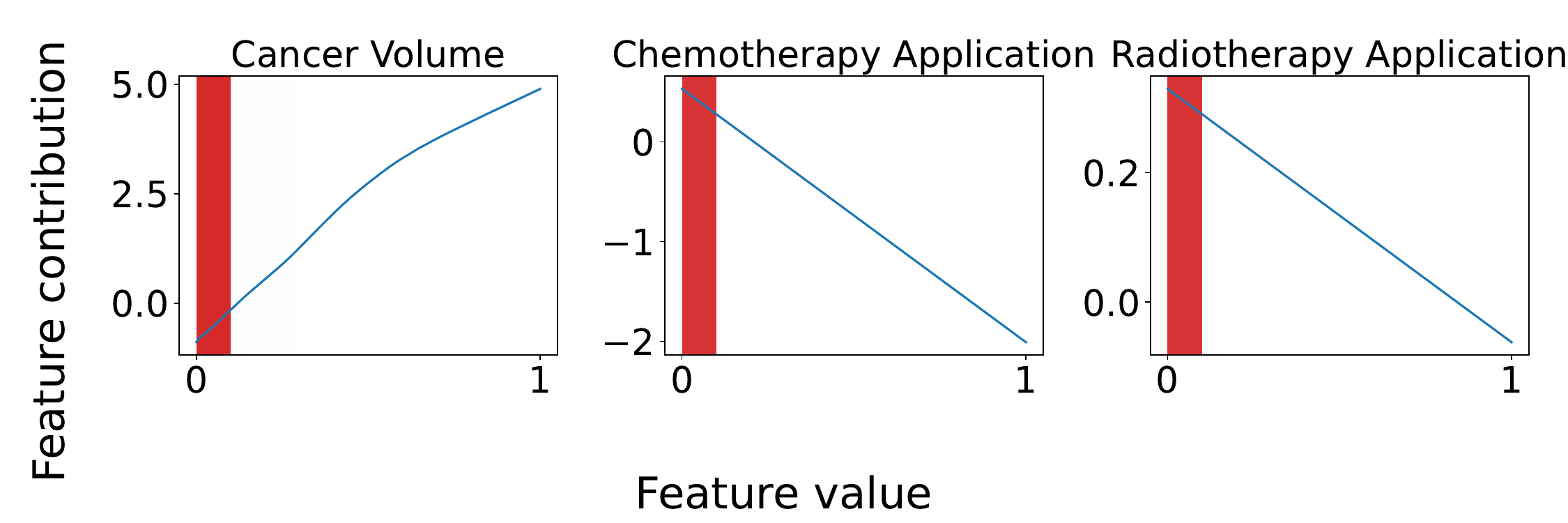}
        \caption{Global interpretations of features in the Cancer dataset.}
        \label{fig:global_cancer}
    \end{minipage}
\end{figure*}

\subsection{Synthetic tumor size prediction}

We conducted an experiment using the synthetic tumor growth dataset developed by \cite{geng2017prediction}, which simulates temporal changes in tumor size. This dataset is beneficial for verifying that the trained GATSM appropriately captures the influence of input features. Tumor size at any given time step is determined by three factors: prior tumor size, chemotherapy, and radiotherapy. For this experiment, we set the coefficient of chemotherapy to 10 and the coefficient of radiotherapy to 1; that is, chemotherapy more significantly reduces tumor size than radiotherapy. The experimental results in Table~\ref{tab:cancer} showed that although GATSM underperforms compared to the complex black-box forecasting models, it nonetheless achieves strong predictive accuracy ($R^2 > 0.9$). Figure~\ref{fig:attention_cancer} and \ref{fig:global_cancer} illustrate the importance of time steps, and the contributions of the three factors. The importance highlights that recent time steps are notably more influential than earlier time steps due to the direct impact of prior tumor size. GATSM also effectively captured the effects of chemotherapy and radiotherapy on reducing tumor size, with chemotherapy having a significantly greater impact than radiotherapy (the scale of the y-axis in chemotherapy is greater than radiotherapy).

\section{Visualizations}\label{sec:vis}

In this section, we provide the visualizations of GATSM's predictions for the AirQuality, Rainfall, and Mortality datasets.

\subsection{AirQuality}

\subsubsection{Time-step importance}
\begin{figure*}[!t]
    \centering
    \includegraphics[width=0.4\linewidth]{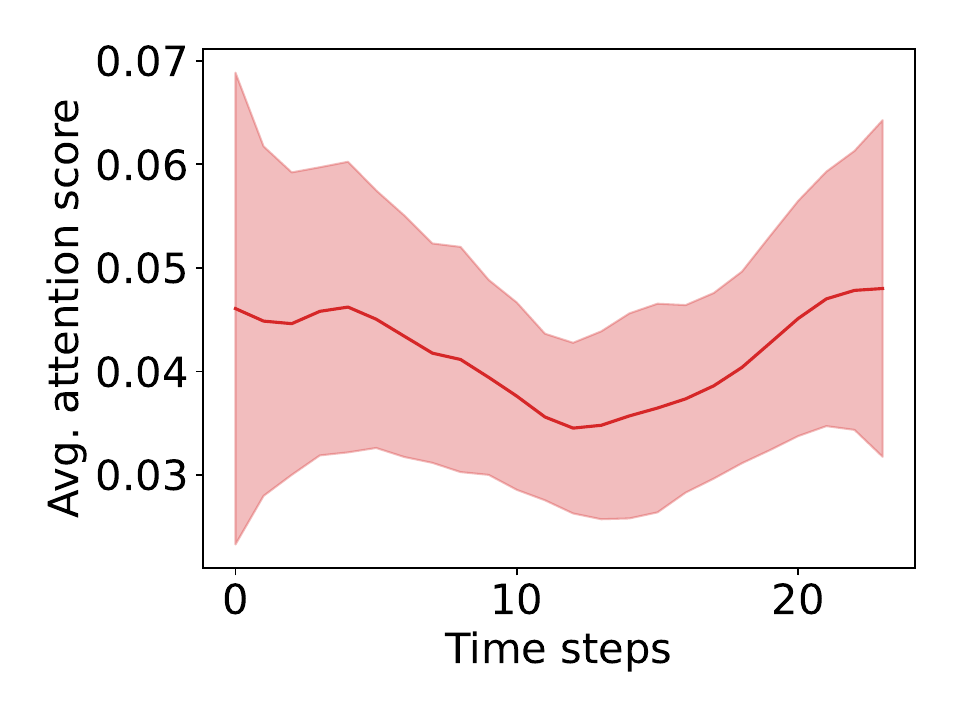}
    \caption{Average attention scores of time steps on the AirQuality dataset.}
        \label{fig:attention}
\end{figure*}

We plot the average attention scores at the last time step $T$ in Figure~\ref{fig:attention}. The process for extracting the average attention score of time step $u$ at time step $t$ is formalized as $\sum_{k=1}^{K} a_{k,t,u}$. This process is repeated over all data samples, and the results are averaged. Based on Figure~\ref{fig:attention}, it seems that GATSM pays more attention to the initial and last states than to the intermediate states. This indicates that the current concentration of particulate matter depends on the initial and recent states.

\subsubsection{Global feature contribution}
Figure~\ref{fig:global} illustrates the global behavior of features in the AirQuality dataset, with red bars indicating the density of training samples. We extract $\sum_{k=1}^{K} h_b \left( x_{t,m} \right) w_{m,b}^{nbm} w_{k,m}^{out}$ from GATSM and repeat this process over the range of minimum to maximum feature values to plot the line. We found that the behavior of \textit{SO2}, \textit{O3}, and \textit{windspeed} is inconsistent with prior human knowledge. Typically, high levels of \textit{SO2} and \textit{O3} are associated with poor air quality. However, GATSM learned that particulate matter concentration starts to decrease when \textit{SO2} exceeds 10 and \textit{O3} exceeds 5. This discrepancy may be due to sparse training samples in these regions, leading to insufficient training, or there may be interactions with other features. Another known fact is that high \textit{windspeed} decreases particulate matter concentration. This is consistent when \textit{windspeed} is below 0.7 in our observation. However, particulate matter concentration drastically increases when \textit{windspeed} exceeds 0.7, likely due to the wind causing yellow dust.

\subsubsection{Local time-independent feature contribution}
\begin{figure*}[!t]
    \centering
    \includegraphics[width=\linewidth]{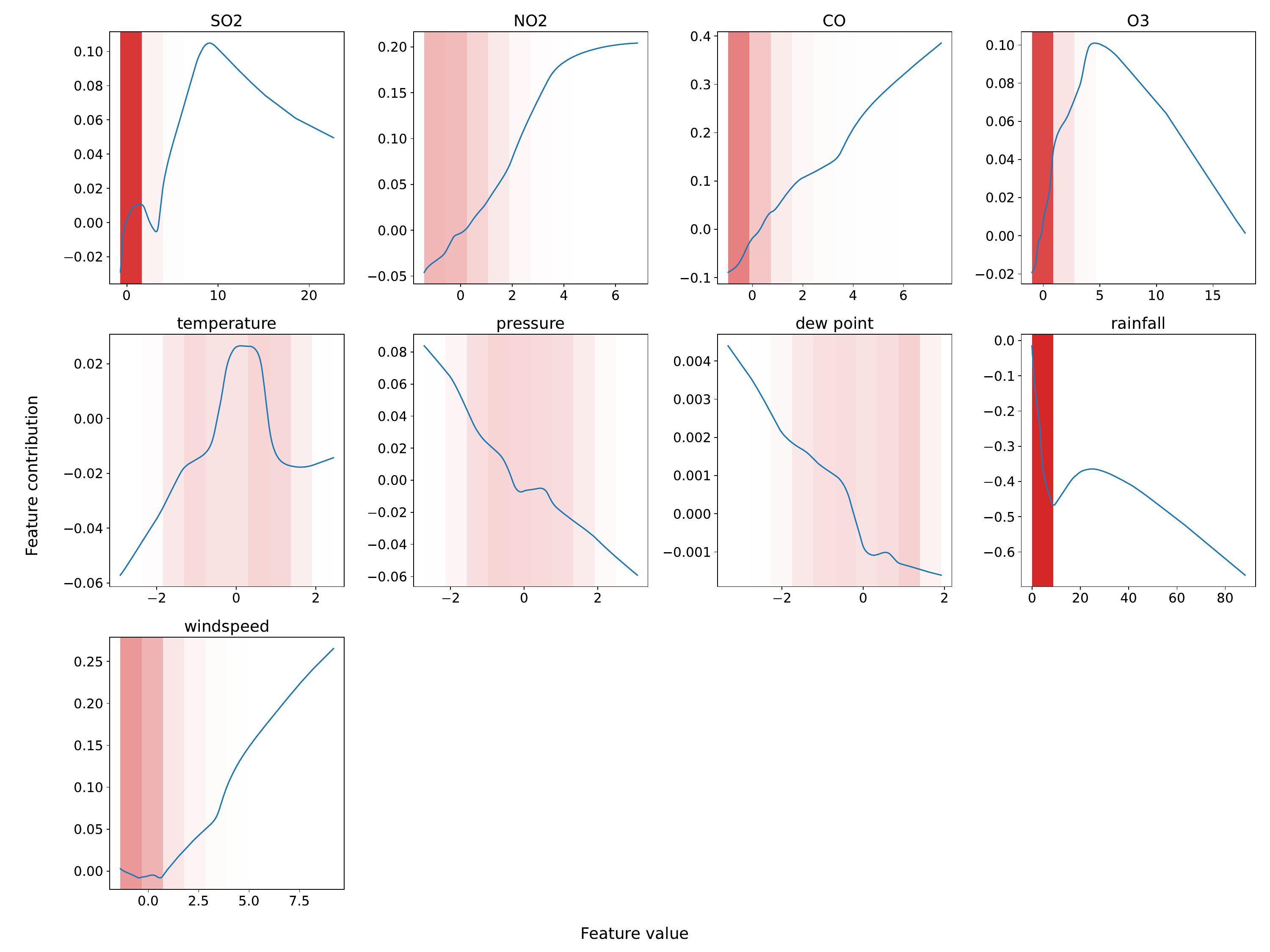}
    \caption{Global interpretations of features in the Air Quality dataset.}
    \label{fig:global}
\end{figure*}

To interpret the prediction of a data sample, we plot the local time-independent feature contributions, $\sum_{k=1}^{K} h_b \left( x_{t,m} \right) w_{m,b}^{nbm} w_{k,m}^{out}$, in Figure~\ref{fig:local1}. The main x-axis (\textcolor{blue}{blue}) represents feature contribution, the sub x-axis (\textcolor{red}{red}) represents feature value, and the y-axis represents time steps. We found that \textit{SO2}, \textit{NO2}, \textit{CO}, and \textit{O3} have positive correlations. In contrast, \textit{temperature}, \textit{pressure}, \textit{dew point}, and \textit{windspeed} have negative correlations. These are consistent with the global interpretations shown in Figure~\ref{fig:global}. Rainfall has the same values across all time steps.

\subsubsection{Local time-dependent feature contribution}
\begin{figure*}[!t]
    \centering
    \includegraphics[width=\linewidth]{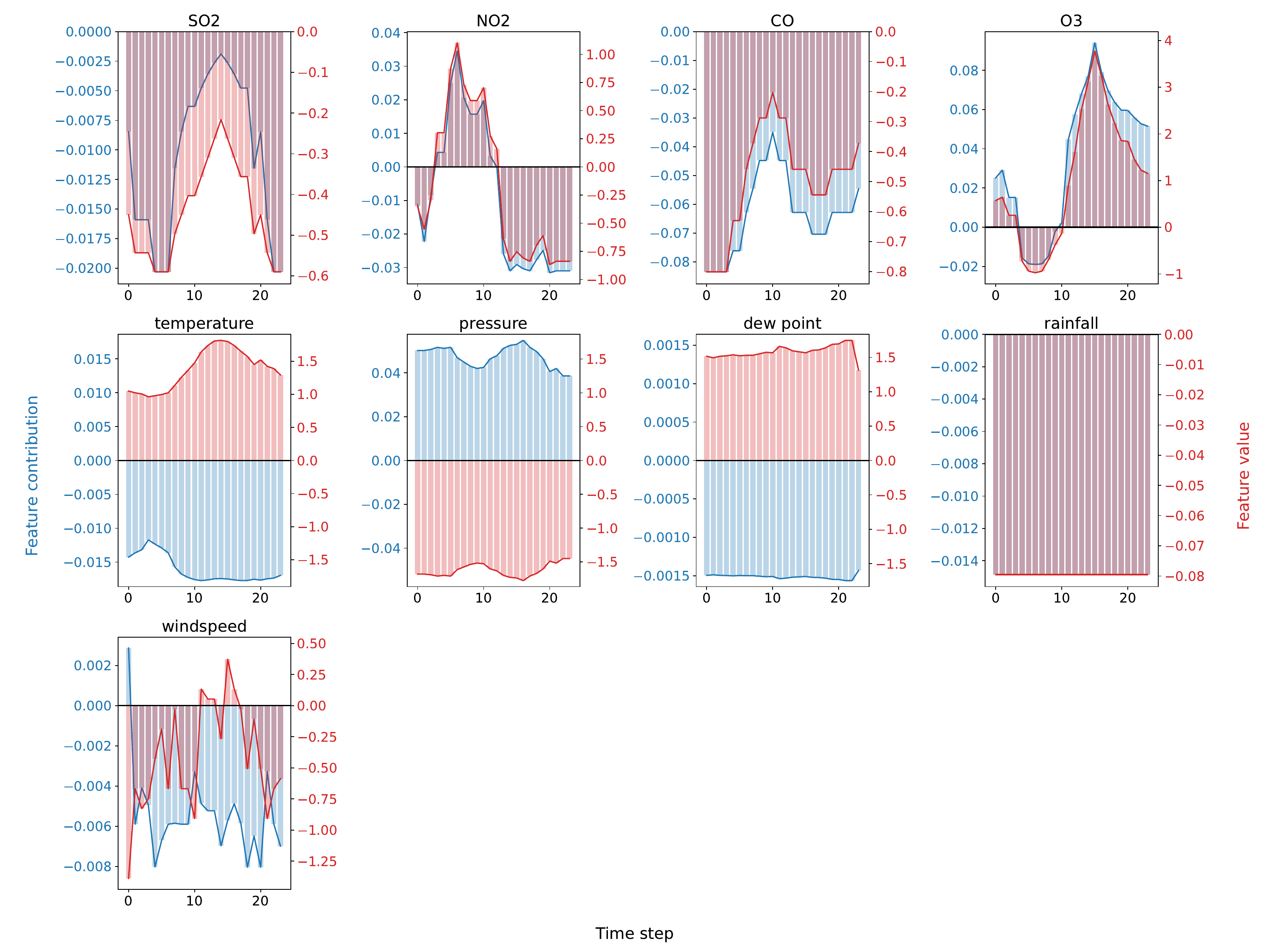}
    \caption{Local time-independent feature contributions.}
    \label{fig:local1}
\end{figure*}

\begin{figure*}[!t]
    \centering
    \includegraphics[width=\linewidth]{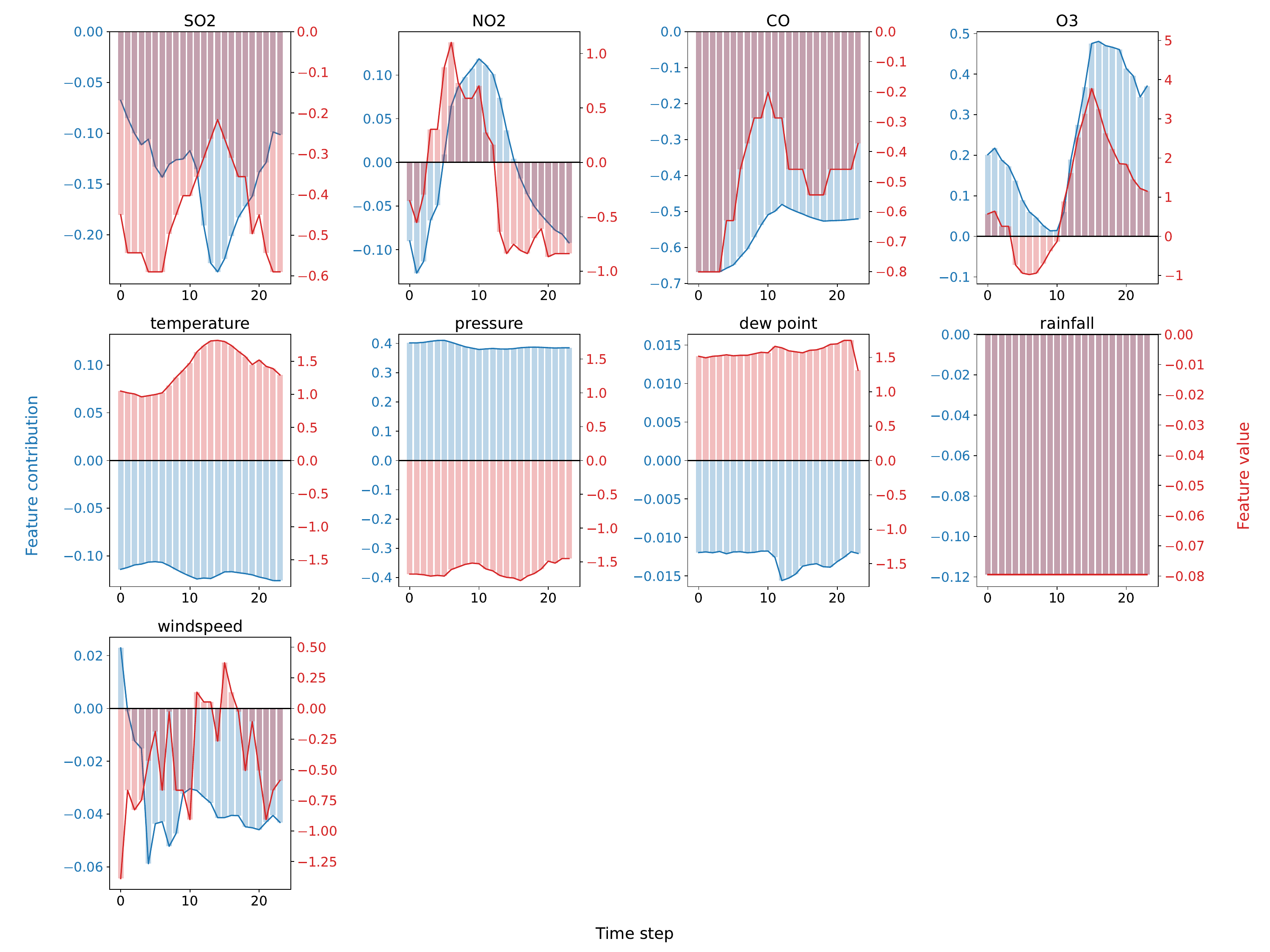}
    \caption{Local time-dependent feature contributions.}
    \label{fig:local2}
\end{figure*}

We also visualize the local time-dependent feature contributions, $\sum_{k=1}^{K} a_{k,t,u} h_b \left( x_{t,m} \right) w_{m,b}^{nbm} w_{k,m}^{out}$. Figure~\ref{fig:local2} illustrates the interpretation of the same data sample as in Figure~\ref{fig:local1}. The time-dependent interpretation differs slightly from the time-independent interpretation. We found that there are time lags in \textit{SO2}, \textit{NO2}, \textit{CO}, and \textit{O3}, meaning previous feature values affect current feature contributions. For example, in the case of \textit{SO2}, low feature values around time step 5 lead to low feature contributions around time step 13.

\subsection{Rainfall}

\subsubsection{Time-step importance}
Figure~\ref{fig:attention_rainfall} illustrates the average importance of all time steps at the final time step. The importance exhibit a cyclical pattern of rising and falling at regular intervals, indicating that GATSM effectively captures seasonal patterns in the Rainfall dataset.

\subsubsection{Global feature contribution:}
Figure~\ref{fig:global_rainfall} illustrates the global behavior of features in the Rainfall dataset, with red bars indicating the density of training samples. Our findings indicate that low \textit{Max Temperature} and high \textit{Min Temperature} contribute to an increase in rainfall.

\subsubsection{Local time-independent feature contribution:}
Figure~\ref{fig:local_rainfall1} shows the local time-independent feature contributions. Consistent with the global interpretation, \textit{Avg. Temperature} and \textit{Min Temperature} have positive correlations with rainfall, while \textit{Max Temperature} has a negative correlation with rainfall.

\subsubsection{Local time-dependent feature contribution:}
\begin{figure*}
        
    \centering
    
    \begin{minipage}{\linewidth}
        \centering
        \includegraphics[width=.35\linewidth]{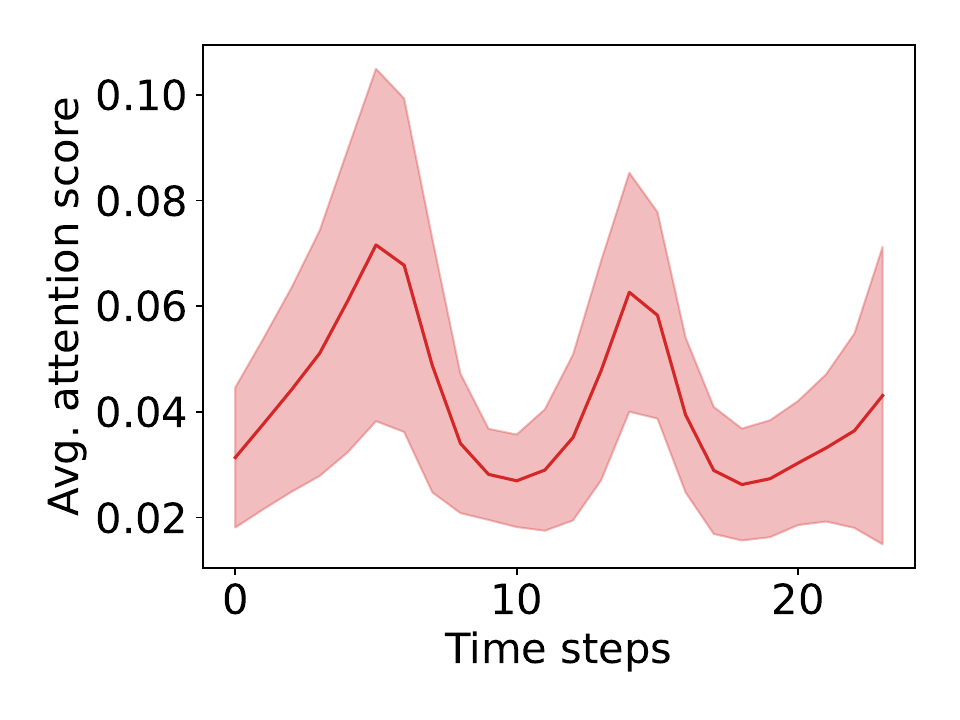}
        \caption{Average attention scores of time steps on the Rainfall dataset.}
        \label{fig:attention_rainfall}
    \end{minipage}
    \begin{minipage}{\linewidth}
        \centering
        \includegraphics[width=.8\linewidth]{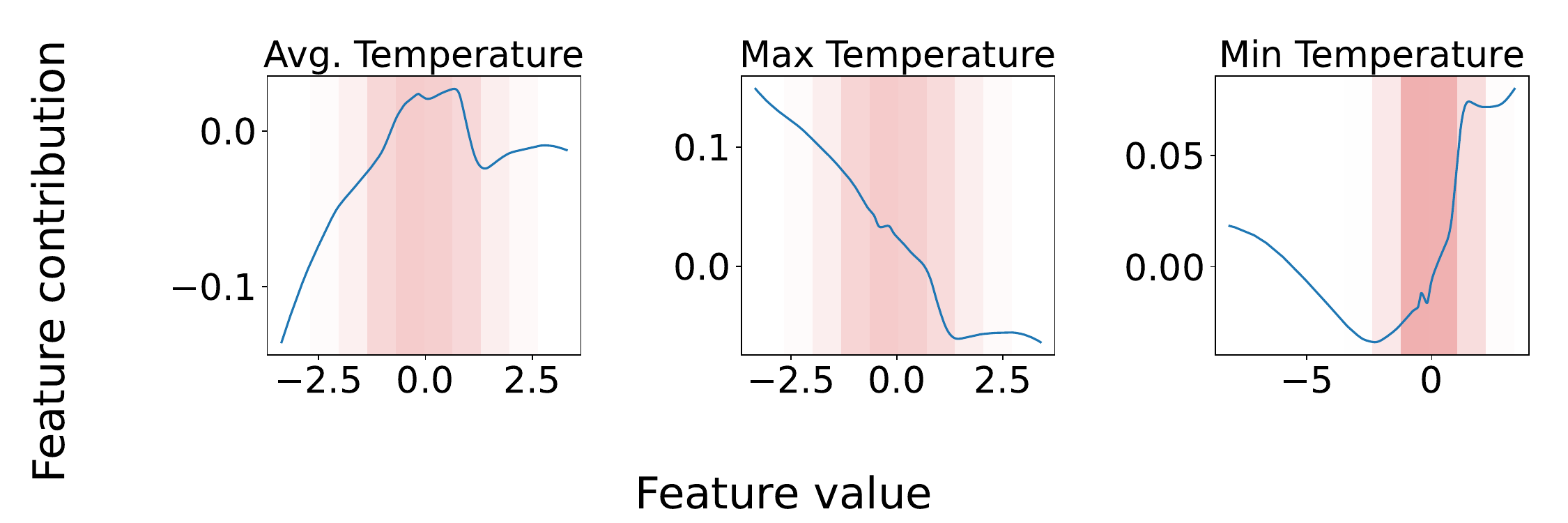}
        \caption{Global interpretations of features in the Rainfall dataset.}
        \label{fig:global_rainfall}
    \end{minipage}
    \begin{minipage}{\linewidth}
        \centering
        \includegraphics[width=\linewidth]{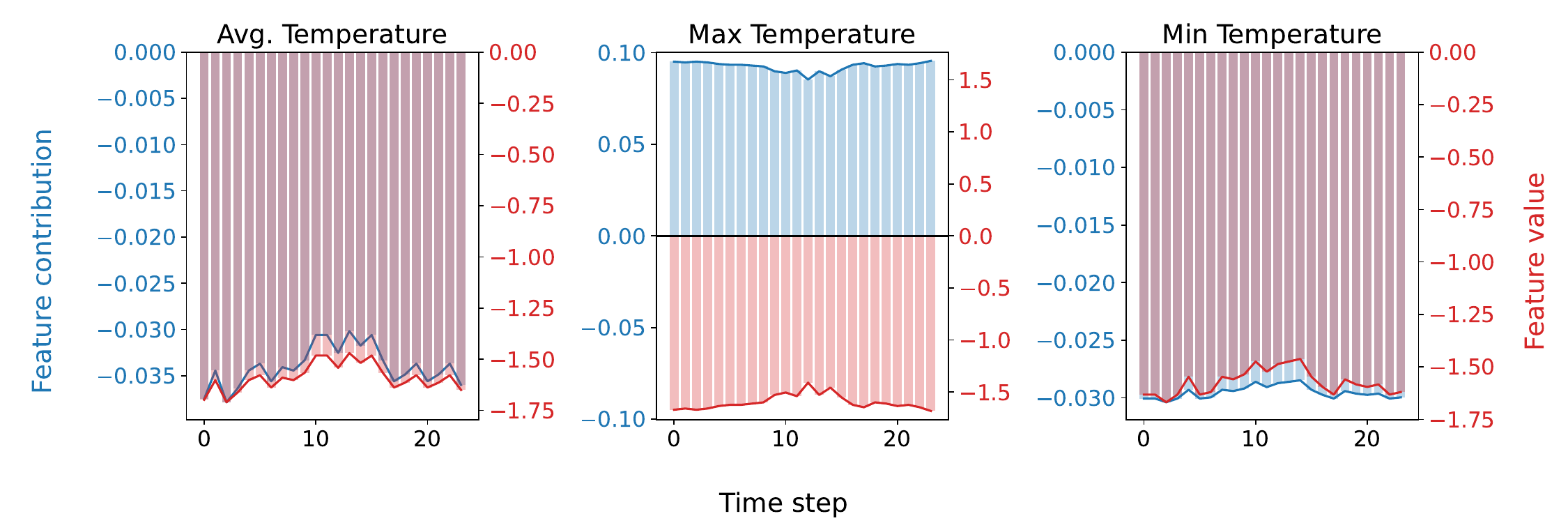}
        \caption{Local time-independent contributions of features in the Rainfall dataset.}
        \label{fig:local_rainfall1}
    \end{minipage}
    \begin{minipage}{\linewidth}
        \centering
        \includegraphics[width=\linewidth]{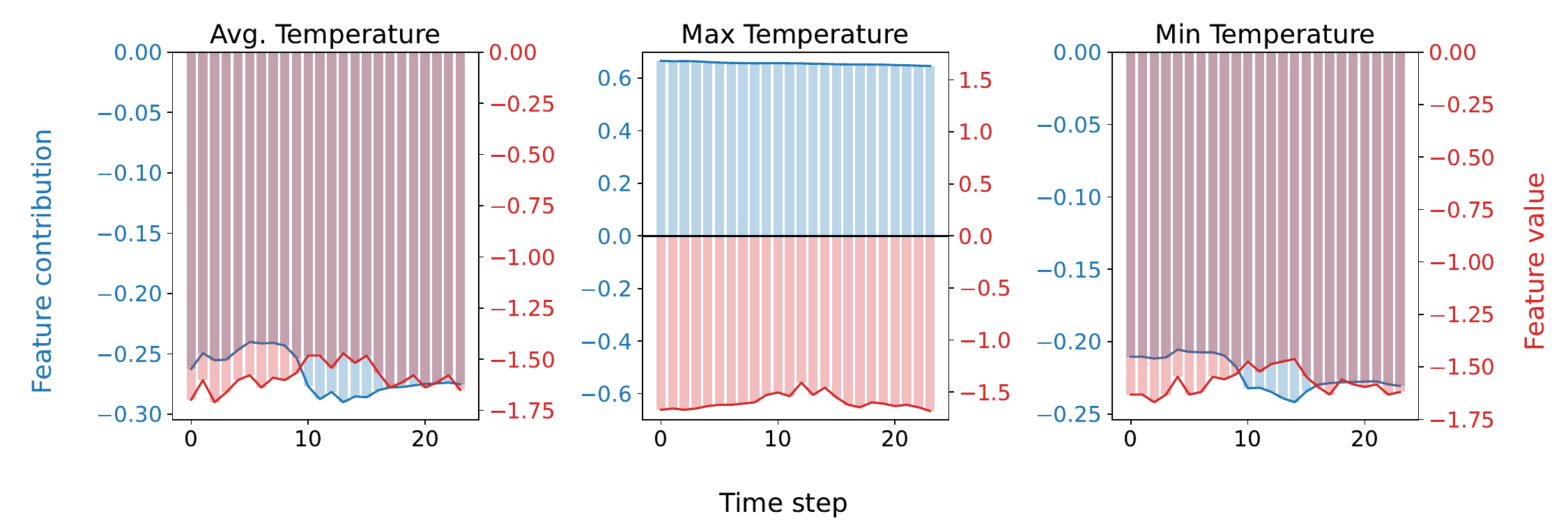}
        \caption{Local time-dependent contributions of features in the Rainfall dataset.}
        \label{fig:local_rainfall2}
    \end{minipage}

\end{figure*}

Figure~\ref{fig:local_rainfall2} shows the local time-dependent feature contributions. All features exhibit patterns similar to the local time-independent contributions. However, we found that \textit{Avg. Temperature} and \textit{Min Temperature} have time lags between feature values and contributions.

\subsection{Mortality}
\begin{figure*}
    \centering
    \includegraphics[width=.9\textwidth]{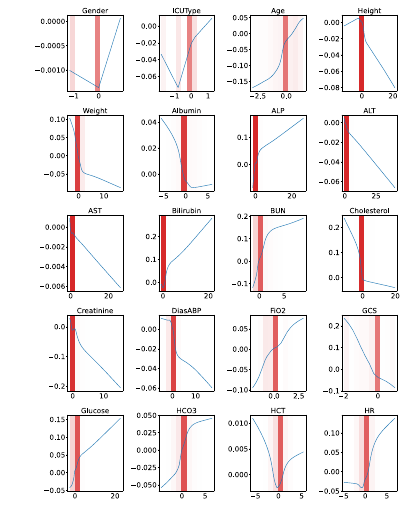}
    \caption{Global interpretations of features in the Mortality dataset (top).}
    \label{fig:global_mortality1}
\end{figure*}

\begin{figure*}
    \centering
    \includegraphics[width=.9\textwidth]{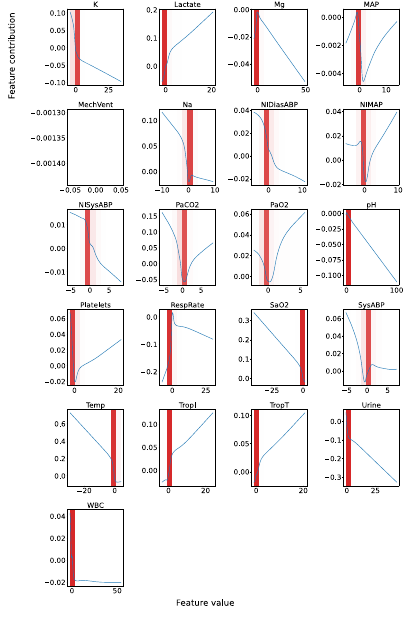}
    \caption{Global interpretations of features in the Mortality dataset (bottom).}
    \label{fig:global_mortality2}
\end{figure*}

\begin{figure*}
    \centering
    \includegraphics[width=.9\textwidth]{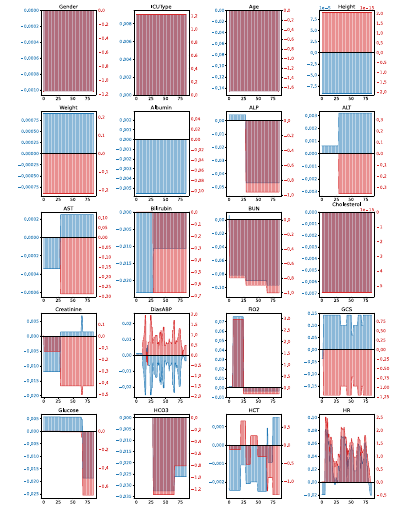}
    \caption{Local time-independent interpretations of features in the Mortality dataset (top).}
    \label{fig:local_mortality11}
\end{figure*}

\begin{figure*}
    \centering
    \includegraphics[width=.9\textwidth]{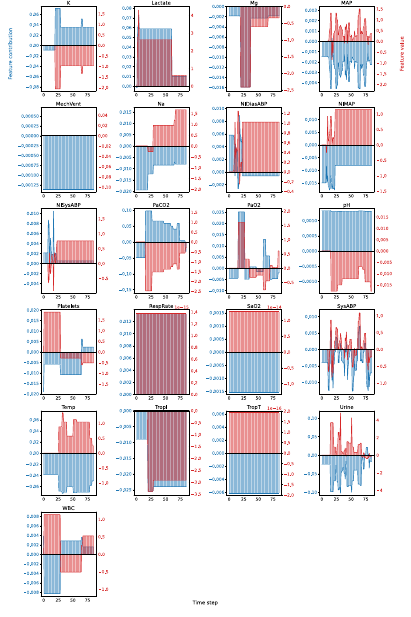}
    \caption{Local time-independent interpretations of features in the Mortality dataset (bottom).}
    \label{fig:local_mortality12}
\end{figure*}

\begin{figure*}
    \centering
    \includegraphics[width=.9\textwidth]{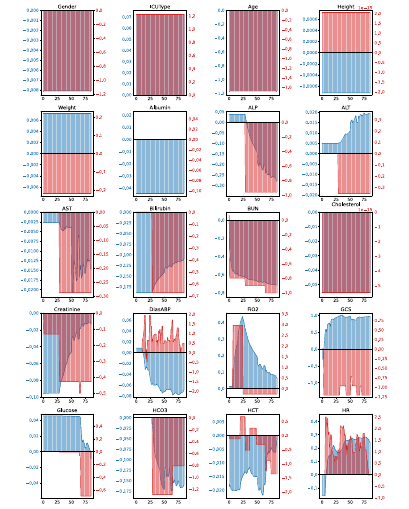}
    \caption{Local time-dependent interpretations of features in the Mortality dataset (top).}
    \label{fig:local_mortality21}
\end{figure*}

\begin{figure*}
    \centering
    \includegraphics[width=.9\textwidth]{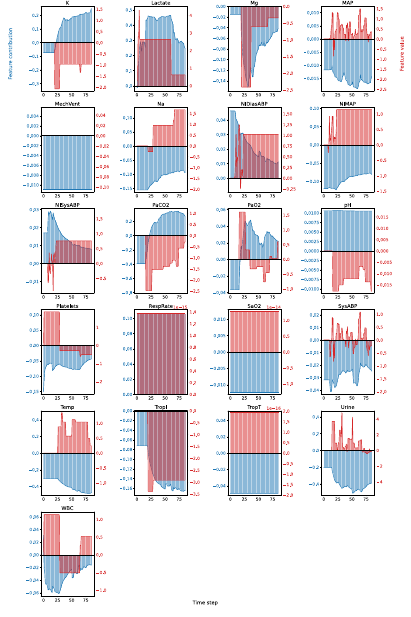}
    \caption{Local time-dependent interpretations of features in the Mortality dataset (bottom).}
    \label{fig:local_mortality22}
\end{figure*}

Figure~\ref{fig:global_mortality1} and Figure~\ref{fig:global_mortality2} illustrates the global behavior of features in the Mortality dataset. Figure~\ref{fig:local_mortality11} and Figure~\ref{fig:local_mortality12} shows the local time-independent feature contributions to the output value. Figure~\ref{fig:local_mortality21} and Figure~\ref{fig:local_mortality22} shows the local time-dependent feature contributions to the output value. Given that the Mortality dataset consists of varying-length time series, the significance of individual time steps differs considerably for each data point. Consequently, we opted not to plot the average importance of time steps.

\clearpage


\clearpage

\bibliography{aaai2026}

\end{document}